\documentclass[journal]{IEEEtran}
\usepackage{cite}
\usepackage{textcomp}
\usepackage{stfloats}
\usepackage{verbatim}
\usepackage{hyperref}
\usepackage{url}

\usepackage[caption=false,font=normalsize,labelfont=sf,textfont=sf]{subfig}
\usepackage{array}
\usepackage{graphicx}
\usepackage{adjustbox} 
\usepackage[dvipsnames]{xcolor}
\usepackage{tabularx}
\usepackage{booktabs}
\usepackage{amsmath,amssymb}
\usepackage{amsthm}
\usepackage{cancel}
\usepackage{mathrsfs}
\usepackage{gensymb}
\usepackage{cases}
\usepackage{mathtools}
\usepackage{gensymb}
\usepackage{bm}
\hypersetup{colorlinks=true,urlcolor=black,linkcolor=black,citecolor=black}
\usepackage{cleveref}
\usepackage{dsfont}
\usepackage{numprint} 
\npthousandsep{,}
\usepackage{marginnote}
\usepackage{mathtools}
\usepackage{multirow}
\usepackage{multicol}
\usepackage{numprint}
\usepackage{longtable}
\usepackage[switch]{lineno}
\usepackage{yfonts}
\usepackage{eufrak}
\usepackage{ulem}
\usepackage{siunitx}
\usepackage{rotating}
\usepackage{algorithm}%
\usepackage{algorithmicx}%
\usepackage{algpseudocode}%
\usepackage{xparse}
\usepackage{comment}
\usepackage{booktabs}
\usepackage{float}
\usepackage[section]{placeins}
\usepackage{microtype}

\NewDocumentCommand{\secondpartialder}{m m o}{\IfNoValueTF{#3}{
		\frac{\partial^2 #1}{\partial #2^2}
	}{
		\frac{\partial^2 #1}{\partial #2 \partial #3}
	}
}

\newcommand{\loss}[0]{\mathscr{L}}

\makeatletter
\newcommand{\slantedparallel}{\mathrel{\mathpalette\new@parallel\relax}}
\newcommand{\new@parallel}[2]{
	\begingroup
	\sbox\z@{$#1T$}%
	\resizebox{!}{\ht\z@}{\raisebox{\depth}{$\m@th#1/\mkern-5mu/$}}%
	\endgroup
}
\makeatother

\newcommand{\eps}[0]{\varepsilon}

\newcommand{\real}[0]{\text{Re}}
\newcommand{\imag}[0]{\text{Im}}
\newcommand{\RR}[0]{\mathbb{R}}
\newcommand{\CC}[0]{\mathbb{C}}
\newcommand{\NN}[0]{\mathbb{N}}

\newcommand{\bfu}[0]{\boldsymbol{u}}
\newcommand{\bfsigma}[0]{\boldsymbol{\sigma}}

\newcommand{\boundarylapl}[1][]{\mathcal{B}_{{\scriptscriptstyle L}\if\relax\detokenize{#1}\relax\else,#1\fi}}
\newcommand{\boundarymech}[1][]{\mathcal{B}_{{\scriptscriptstyle M}\if\relax\detokenize{#1}\relax\else,#1\fi}
}

\theoremstyle{definition}

\crefname{theorem}{theorem}{theorems}
\Crefname{theorem}{Theorem}{Theorems}

\crefname{lemma}{lemma}{lemmas}
\Crefname{lemma}{Lemma}{Lemmas}

\crefname{proposition}{proposition}{propositions}
\Crefname{proposition}{Proposition}{Propositions}

\crefname{problem}{problem}{problems}
\Crefname{problem}{Problem}{Problems}

\crefname{definition}{definition}{definitions}
\Crefname{definition}{Definition}{Definitions}

\crefname{remark}{remark}{remarks}
\Crefname{remark}{Remark}{Remarks}

\crefname{observation}{observation}{observations}
\Crefname{observation}{Observation}{Observations}

\crefname{example}{example}{examples}
\Crefname{example}{Example}{Examples}

\definecolor{raspberry}{rgb}{1,0,0.2}
\definecolor{darkred}{rgb}{0.5,0.1,0.1}
\definecolor{darkgreen}{rgb}{0.1,0.6,0.2}

\title{AURA: Angular Update Rate Adaptation for training complex-valued neural networks}

\author{
	Enrico~Ballini\textsuperscript{1},
	Allan~Peter~Engsig-Karup\textsuperscript{2},
	and~Tito~Andriollo\textsuperscript{1}
	\thanks{Corresponding author: Enrico Ballini (e-mail: enrico.ballini@gmail.com).}
	\thanks{\textsuperscript{1}Aarhus University, Aarhus, Denmark.}
	\thanks{\textsuperscript{2}Technical University of Denmark, Kongens Lyngby, Denmark.}
}

\begin{document}
\bstctlcite{IEEEexample:BSTcontrol}

\maketitle

\begin{abstract}
	Complex-valued neural networks (CVNNs) are increasingly adopted for complex-valued data; however, they are often trained with first-order optimizers inherited from the real-valued case.
	The efficiency of these methods depends largely on the step size, and their step-size rules ignore the angular information available in the complex plane.
	We address step-size adaptation in the complex domain by introducing AURA (Angular Update Rate Adaptation), a per-parameter step-size adaptation that can be added on top of any first-order optimizer, and removed from it, without altering its update direction.
	AURA measures the agreement between consecutive updates of each complex parameter, in length, alignment, and sense of rotation, and enlarges the step when they are consistent and reduces it when they are not.
	It requires no additional gradient evaluations and only inexpensive vector operations per step.
	We combine AURA with Adam and Muon and compare the resulting methods with well-known first-order optimizers on four test cases of increasing complexity, ranging from the approximation of scalar complex functions to physics-informed training. Fully connected neural networks are used throughout this work. All hyperparameters other than the step size are held fixed across test cases; for one case, we also tune the hyperparameters of each optimizer under the same budget.
	Our empirical tests show that AURA improves the convergence of its base optimizer in most cases with a small per-step overhead, and we identify the conditions under which it fails to do so.
\end{abstract}


\section{Introduction}\label{sec:introduction}

Neural networks are trained with iterative optimizers, ranging from first- to second-order methods and from heuristic rules to principled ones \cite{Bottou2018}.
First-order methods are the standard in deep learning because each step is cheap, but their efficiency depends largely on the step size.
The classical levers are the gradient history and its reliability: momentum lets noise cancel and consistent directions accumulate, while the signal-to-noise ratio (SNR) of the gradient decides how far a step can be trusted, as in AdaBelief \cite{Zhuang2020}.

A further device to adjust the step size is the agreement between consecutive gradients or updates; for real-valued networks, AngularGrad \cite{Roy2021} rescales the Adam step with an angle-like measure of consecutive gradients, diffGrad \cite{Dubey2020} with their difference, and AdaSmooth \cite{Lu2023} sets the averaging window of the RMSProp denominator from the ratio between the net and the total displacement of a weight.
Hypergradient descent \cite{Gunes2018}, applied earlier to complex-valued adaptive filters \cite{Goh2005, Goh2007}, learns the step size itself by gradient descent, and its multiplicative form updates the step size with the cosine between the current gradient and the previous update.
Adaptive methods such as AdaGrad \cite{Duchi2011} and AdaDelta \cite{Zeiler2012} keep one step size per parameter, and CoRe \cite{Eckhoff2023, Eckhoff2024} shows that adding an RPROP-like sign test to Adam at this level can be effective in real-valued networks.

Whether such rules actually speed up training can only be judged through fair comparisons.
Benchmarking practices are set in, e.g., \cite{Schneider2018} and \cite{Schmidt2021}, based on standard test problems and explicit tuning budgets, and take into account that the hyperparameter search space can change the ranking \cite{Choi2020}.

The above works focused on the real-valued case, see also \cite{Kiyani2025}. For complex-valued neural networks (CVNNs), \cite{Mayer2025} compare adaptive optimizers in online training only, and \cite{Rutkowski2026} show that the specific optimizer changes how CVNNs compare with real-valued neural networks (RVNNs).

CVNNs process complex-valued data directly and are used in image classification \cite{Trabelsi2018}, image reconstruction, signal classification, speech enhancement, wind prediction, and control systems \cite{Lee2022}.
They are a natural choice in telecommunications \cite{Mayer2025}, and in MRI reconstruction they outperform real-valued CNNs with the same number of parameters \cite{Cole2021}.
Their advantage is not universal, however: \cite{Almansoori2025} find RVNNs as good or better in most cases, and the optimizer is part of the reason \cite{Rutkowski2026}.

Several training algorithms have been designed for CVNNs, from metacognitive learning in complex RBF networks \cite{Savitha2012a} to conjugate gradients and selectable search directions with Armijo \cite{Zhang2019} and Wolfe \cite{Dong2021} line searches.
Adaptive optimizers, instead, have been ported from the real-valued case by treating the real and imaginary parts separately: \cite{Zhang2019a} extend Adam in this way, and \cite{Mayer2025} extend AdaGrad \cite{Duchi2011}, RMSProp \cite{Hinton2012}, AdaMax \cite{Kingma2014}, AMSGrad \cite{Sashank2018}, SAMSGrad \cite{Tong2022}, Nadam \cite{Dozat2016}, and diffGrad \cite{Dubey2020}.
Their step-size rules therefore act on each part independently and measure no angle in the complex plane.

The complex plane also allows a complex step size, which rotates the update and extends the search from a ray to a half-plane: \cite{Zhang2016} show that the complex step size approximates Hessian information better than a real one, and \cite{Zhang2020} and \cite{Zhao2023} adapt it during training.
The authors of \cite{Yin2026} propose to adjust a complex step size, shared by all parameters, according to its phase, obtained by comparing gradient norms at three trial points; the angle thus comes from flatness rather than from consecutive updates, at the cost of three extra gradient evaluations per step.
Methods based on curvature or line searches are also more expensive per step: HCSCGM \cite{Zhang2024} and its extension with an adaptive complex step size, HCSCGACS \cite{Zhang2025a}, rely on complex inexact line searches, AS-CNAG \cite{Zhao2024} sets the step from an approximate Hessian, and \cite{Zhang2023, Wang2023a} develop adaptive complex L-BFGS methods.
We therefore focus on optimizers that use one stochastic gradient per step and inexpensive vector operations.

This work builds on the idea that the angle between consecutive updates is an informative signal for step-size control in CVNNs.
We propose AURA, a per-parameter step-size multiplier applicable on top of any optimizer, which enlarges the step when consecutive update directions are aligned and of similar length and reduces it when they are opposed or keep rotating in the same sense.
We combine AURA with Adam \cite{Kingma2014} and Muon \cite{Jordan2024} and evaluate it on four problems with fully connected neural networks, from scalar complex functions to more complicated physics-informed training.
Unlike real-valued rules, which can only test whether consecutive updates agree in sign and magnitude, AURA also measures how much they rotate in the complex plane, in which sense they rotate, and how similar they are in magnitude. The building blocks of AURA have already been used in the cited literature, but their combination for complex parameters is new.
\Cref{subsec:aura} presents the AURA algorithm, \Cref{sec:related_adaptive_optimizers} relates it to the closest optimizers, \Cref{sec:benchmarks} reports the case studies, and \Cref{sec:conclusion} concludes.

\section{AURA}
\label{subsec:aura}

The method we propose scales the update direction of a base optimizer by a per-parameter multiplier, which, in a nutshell, increases when consecutive directions agree in the complex plane and decreases when they do not. This section fixes the notation, states the algorithm, and explains the building blocks of the algorithm.

\paragraph{Notation}
Let $w_{j,t} = x_{j,t} + \mathrm i\,y_{j,t} \in \CC$, $j = 1,\ldots,N$, be the $N$ trainable parameters at iteration $t \in \NN$, and $\loss_t:\CC^N \to \RR$ the loss at step $t$, e.g., the mini-batch loss \eqref{eq:benchmark_batch_loss}.
Being real-valued, $\loss_t$ is not holomorphic and is differentiated as a map $\RR^{2N} \to \RR$ under $\CC^N \simeq \RR^{2N}$. The gradient
\begin{equation}
	g_{j,t} = \frac{\partial\loss_t}{\partial x_{j,t}} + \mathrm i\,\frac{\partial\loss_t}{\partial y_{j,t}},
	\label{eq:aura_gradient}
\end{equation}
combines the two real partial derivatives into a single complex number, and $-g_{j,t}$ is the steepest-descent direction of $\loss_t$ in $w_{j,t}$.

\paragraph{Algorithm}
Given the direction $d_{j,t}$ of a base optimizer $\mathcal{G}$, AURA computes a multiplier $\gamma_{j,t} > 0$ and takes the step $\alpha\gamma_{j,t}d_{j,t}$.
With the initial values $d_{j,0} = 0$, $\tilde\zeta_{j,0} = 0$, and $\gamma_{j,0} = 1$, and suitable choices for the hyperparameters listed in \Cref{tab:aura_hyperparameters} and discussed below, the AURA algorithm takes the following form for every $j$ and $t \geq 1$:
\begin{subequations}\label{eq:aura}
\begin{align}
	\vcenter{\hbox{\rotatebox{90}{\text{\shortstack[c]{Optimizer\\ (Adam, Muon, \ldots)}}}}}& \quad d_{j,t} = \mathcal{G}_j\bigl(\{g_{k,s}\}_{k \leq N,\, s \leq t}\bigr),\label{eq:direction}\\
	\vcenter{\hbox{\rotatebox{90}{\text{\shortstack[c]{Consistency\\ variable}}}}}&\left\{
		\begin{aligned}
			\zeta_{j,t} &=
			\frac{2\,d_{j,t}\overline{d_{j,t-1}}}
			{|d_{j,t}|^2+|d_{j,t-1}|^2+\eps_{\mathrm E}},
			\qquad |\zeta_{j,t}|<1,\\
			\tilde{\zeta}_{j,t} &=
			\beta_\zeta\tilde{\zeta}_{j,t-1}
			+(1-\beta_\zeta)\zeta_{j,t},\qquad
			\widehat{\zeta}_{j,t} =
			\frac{\tilde{\zeta}_{j,t}}{1-\beta_\zeta^{\,t}},\\
			\chi_{j,t} &=
			\real\bigl(\widehat{\zeta}_{j,t}\bigr)\in(-1,1),\\
			\Psi_{j,t} &=
			\imag\bigl(\widehat{\zeta}_{j,t}\bigr)\in(-1,1),
		\end{aligned}
	\right.\label{eq:consistency}\\
	\vcenter{\hbox{\rotatebox{90}{\shortstack[c]{``Delta-bar-delta''-style\\$\alpha$ adjustment}}}}&\left\{
		\begin{aligned}
			\sigma_{j,t} &=
			\bigl(\chi_{j,t}\leq\chi_{\mathrm o}\bigr)
			\ \text{or}\
			\bigl(|\Psi_{j,t}|\geq\Psi_{\mathrm o}\bigr),\\
			\rho_{j,t} &=
			\bigl(\chi_{j,t}\geq\chi_{\mathrm a}\bigr)
			\ \text{and}\
			\bigl(|\Psi_{j,t}|\leq\Psi_{\mathrm a}\bigr),\\
			\gamma_{j,t} &=
			\begin{cases}
				\max\{\eta_-\gamma_{j,t-1},\gamma_{\min}\},
				&\sigma_{j,t} = 1,\\
				\min\{\eta_+\gamma_{j,t-1},\gamma_{\max}\},
				&\rho_{j,t} = 1,\\
				\gamma_{j,t-1},&\text{otherwise},
			\end{cases}
		\end{aligned}
	\right.\label{eq:lr_multiplier}\\
	&w_{j,t+1}
	= w_{j,t}-\alpha\gamma_{j,t}d_{j,t}-\alpha\lambda w_{j,t}.
	\label{eq:weight_update}
\end{align}
\end{subequations}
\paragraph{Base optimizer}
Line \eqref{eq:direction} is any first-order optimizer, e.g., Adam \cite{Kingma2014} or Muon \cite{Jordan2024}; the subsequent blocks depend on $d_{j,t}$ only.
With $\gamma_{j,t} = 1$ for all $j$ and $t$, AURA reduces to $\mathcal{G}$, which shows that the proposed algorithm acts as a plug-in to the base optimizer.

\paragraph{Consistency variable}
Block \eqref{eq:consistency} compares $d_{j,t}$ with $d_{j,t-1}$ through the complex variable $\zeta_{j,t}$, a Dice-like measure \cite{Egghe2009} (\Cref{app:salton_dice}).
Under $\CC \simeq \RR^2$, the real part of its numerator is the inner product of the two directions and the imaginary part their signed area; the denominator normalizes by their squared lengths, so that $\zeta_{j,t} \to 1$ for equal directions, $\zeta_{j,t} \to -1$ for opposite ones, $\zeta_{j,t} \to e^{\mathrm i\theta}$ for a rotation by $\theta$ at equal length, and $|\zeta_{j,t}|$ decreases as the two lengths separate.
Unlike a cosine similarity, $\zeta_{j,t}$ therefore captures length, angle, and sense of rotation simultaneously; the safeguard $\eps_{\mathrm E} > 0$ only prevents division by zero.
The moving average $\tilde\zeta_{j,t}$, with decay rate $\beta_\zeta$, retains the memory of past directions and, since it starts from zero, is bias-corrected.
The real part $\chi_{j,t}$ of $\widehat\zeta_{j,t}$ is the smoothed alignment of consecutive directions, the imaginary part $\Psi_{j,t}$ their smoothed signed rotation: rotations of the same sense accumulate in $\Psi_{j,t}$, alternating ones cancel.

\paragraph{Multiplier and update}
Block \eqref{eq:lr_multiplier} converts the two statistics into $\gamma_{j,t}$ by a three-way test, conceptually similar to the strategies delta-bar-delta \cite{Jacobs1988} and RPROP \cite{Riedmiller1993}: the multiplier is increased by $\eta_+ > 1$ when the flag $\rho_{j,t}$ holds, decreased by $\eta_- \in (0,1)$ when $\sigma_{j,t}$ holds, and left unchanged otherwise, always within $[\gamma_{\min},\gamma_{\max}]$.
Alignment alone is not sufficient for an increase: a sequence of small rotations of constant sense keeps $\chi_{j,t}$ close to $1$ but accumulates in $\Psi_{j,t}$.
An increase therefore requires both $\chi_{j,t} \geq \chi_{\mathrm a}$ and $|\Psi_{j,t}| \leq \Psi_{\mathrm a}$, whereas either $\chi_{j,t} \leq \chi_{\mathrm o}$ or $|\Psi_{j,t}| \geq \Psi_{\mathrm o}$ forces a decrease; the two flags are exclusive since $\chi_{\mathrm o} < \chi_{\mathrm a}$ and $\Psi_{\mathrm a} < \Psi_{\mathrm o}$.
The update \eqref{eq:weight_update} applies the scaled direction with base step size $\alpha > 0$ and decoupled weight decay $\lambda \geq 0$, as in AdamW \cite{Loshchilov2018}.

\paragraph{Hyperparameters and cost}
AURA adds the ten hyperparameters of \Cref{tab:aura_hyperparameters}, plus the weight decay $\lambda$; the numerical values used in the test cases (\Cref{sec:benchmarks}) are given in \Cref{tab:optimizer_settings_shared}.
Most have a clear interpretation and should not require heavy tuning: $\eps_{\mathrm E}$ is a safeguard, $\gamma_{\min}$ and $\gamma_{\max}$ bound the multiplier, and $\eta_\pm$ set its rate of change.
The decay rate $\beta_\zeta$ and the four thresholds can be more difficult to tune.
Per parameter, AURA stores $d_{j,t-1}$, $\tilde\zeta_{j,t}$, and $\gamma_{j,t}$, i.e., five real numbers against the three of Adam (a complex first moment and a real second moment), and adds a few elementwise operations per step; it needs no extra gradient evaluations.
The measured overhead is reported in \Cref{sec:benchmarks}.

\begin{table*}[!t]
	\centering
	\caption{Hyperparameters of AURA}
	\label{tab:aura_hyperparameters}
	\small
	\begin{tabular}{@{}lll@{}}
		\toprule
		Symbol & Range & Role \\
		\midrule
		\multicolumn{3}{@{}l}{\emph{Consistency variable \eqref{eq:consistency}}}\\
		$\beta_\zeta$ & $[0,1)$ & Decay rate of the average $\tilde\zeta_{j,t}$ \\
		$\eps_{\mathrm E}$ & $> 0$ & Safeguard on the denominator of $\zeta_{j,t}$ \\
		\midrule
		\multicolumn{3}{@{}l}{\emph{``Delta-bar-delta''-style $\alpha$ adjustment \eqref{eq:lr_multiplier}}}\\
		$\chi_{\mathrm a}$ & $(\chi_{\mathrm o},1]$ & Alignment threshold for growth \\
		$\chi_{\mathrm o}$ & $[-1,\chi_{\mathrm a})$ & Opposition threshold forcing decrease \\
		$\Psi_{\mathrm a}$ & $[0,\Psi_{\mathrm o})$ & Rotation tolerance for growth \\
		$\Psi_{\mathrm o}$ & $(\Psi_{\mathrm a},1]$ & Rotation threshold forcing decrease \\
		$\eta_-$ & $(0,1)$ & Decrease factor of $\gamma_{j,t}$ \\
		$\eta_+$ & $> 1$ & Increase factor of $\gamma_{j,t}$ \\
		$\gamma_{\min}$ & $(0,1]$ & Floor on the multiplier \\
		$\gamma_{\max}$ & $\geq 1$ & Ceiling on the multiplier \\
		\midrule
		\multicolumn{3}{@{}l}{\emph{Update \eqref{eq:weight_update}}}\\
		$\lambda$ & $\geq 0$ & Decoupled (AdamW) weight decay \\
		\bottomrule
	\end{tabular}
\end{table*}

\section{Related work}
\label{sec:related_adaptive_optimizers}

This section relates AURA to the optimizers it builds on or resembles, with a focus on how each method measures the agreement between successive gradients or updates and on how it uses that measurement.

\paragraph{Adam and Muon}
Adam \cite{Kingma2014} and Muon \cite{Jordan2024} are the base optimizers $\mathcal{G}$ used in this work because they are established methods with known good performance, although other optimizers could be chosen.
Adam normalizes the first moment of the gradient by the square root of its second moment, which Balles and Hennig \cite{Balles2018} interpret as a damping of each step by the signal-to-noise ratio (SNR) of the corresponding gradient component.
Muon applies momentum and orthogonalizes the update of each weight matrix by Newton--Schulz iterations; it carries no per-parameter variance estimate.
In both cases, the SNR does not enter $\zeta_{j,t}$, which depends only on the directions $d_{j,t}$ and $d_{j,t-1}$: AURA inherits whatever variance adaptation the base optimizer provides and adds none of its own.
This could represent a limitation for noisy updates, since $\zeta_{j,t}$ then measures the agreement between noisy directions (see also \Cref{sec:test_case_holomorphic}).

\paragraph{Step-size adaptation by sign information}
The idea of adapting the step size from the agreement of successive derivatives dates back to the rules of Kesten, Saridis, and Barto and Sutton \cite{Kesten1958, Saridis1970, BartoSutton1981}, and was systematized by Jacobs \cite{Jacobs1988} in the delta-bar-delta rule, which increases the step size of a weight while consecutive derivatives share their sign and decreases it when they differ.
AURA rests on the same principle, with the sign test replaced by a test on the complex quantity $\widehat\zeta_{j,t}$.

\paragraph{RPROP}
The method proposed by \cite{Riedmiller1993, Riedmiller1994} assigns each weight its own update value, which is increased by a factor $\eta_+ > 1$ while the partial derivative keeps its sign and decreased by a factor $\eta_- < 1$ when the sign changes; the weight then moves by this value against the sign of the derivative, irrespective of its magnitude.
As in AURA, the agreement of successive derivatives sets the size of the next step.
Complex RPROP \cite{Kantsila2004, Popa2014} applies the same rule to the real and imaginary parts separately, and therefore measures no angle in the complex plane.

\paragraph{CoRe}
The CoRe optimizer \cite{Eckhoff2023, Eckhoff2024} combines an Adam direction $u_{j,t} \in \RR$, the counterpart of $d_{j,t}$ in \eqref{eq:direction}, with an RPROP-style per-weight step size $s_{j,t} \in \RR$, as AURA does, and differs from it in three respects.
First, it is defined for real-valued networks: its step-size rule depends only on the sign of the product of two consecutive first moments, the real-domain counterpart of the sign of $\chi_{j,t}$.
Second, the rule is gated by a binary plasticity factor $P_{j,t} \in \{0,1\}$, which freezes the weights with a high importance score. The update reads
\begin{equation}\label{eq:related_core_update}
	w_{j,t}  = w_{j,t-1} - \underbrace{u_{j,t}\,P_{j,t}\,s_{j,t}}_{\text{update}} - \underbrace{d_\chi\,|u_{j,t}|\,P_{j,t}\,s_{j,t}w_{j,t-1}}_{\text{regularization}},
\end{equation}
where $d_\chi \in \RR_{\geq 0}$ is a hyperparameter.
Since $P_{j,t}$ also multiplies the update, a weight can remain unchanged for a whole step even when $u_{j,t}$ is nonzero.
AURA has no such switch: $\gamma_{j,t}$ is bounded below by $\gamma_{\min} > 0$, so the step in \eqref{eq:weight_update} is never suppressed while $d_{j,t}$ is nonzero.
Third, the regularization term in \eqref{eq:related_core_update} is a weight decay proportional to the magnitude $|u_{j,t}|P_{j,t}s_{j,t}$ of the update, with a group-specific coefficient $d_\chi$, so that stable weights are decayed the least.
Overfitting control is not a target of AURA, which uses the decoupled weight decay of Loshchilov and Hutter \cite{Loshchilov2018} and is compatible with other strategies.

\paragraph{AdaRem}
AdaRem \cite{Liu2020} multiplies the step size of each parameter by $a_{t,i} = 1 + \lambda_t b_{t,i}$, where $b_{t,i} = g_{t,i}m_{t,i}/(|g_{t,i}|\max|m_t| + \epsilon)$ and $m_t$ is the exponential moving average of the gradient.
Up to $\epsilon$, $b_{t,i} = \mathrm{sign}(g_{t,i}m_{t,i})\,|m_{t,i}|/\max|m_t|$: for two scalars, Salton's cosine reduces to the sign alone, which AdaRem weights by the relative size of the momentum.
As in AURA, the step increases when the current gradient agrees with the past direction and decreases when it opposes it.

\paragraph{TAM}
The algorithm TAM \cite{Malviya2024} computes the cosine similarity (Salton's measure, \Cref{app:salton_dice}) between the previous momentum direction and the current gradient, an ingredient close to $\chi_{j,t}$; this quantity, however, modifies the momentum rather than the step size.

\paragraph{diffGrad}
diffGrad \cite{Dubey2020} multiplies the Adam direction by a ``friction coefficient'' $\xi_{j,t} = \bigl(1+e^{-|g_{j,t}-g_{j,t-1}|}\bigr)^{-1} \in [\tfrac12,1)$, so that ``large changes in the gradient incur less friction, whereas small changes in the gradient incur more friction'' \cite{Dubey2020}.
Its behavior is therefore, to some extent, opposite to that of AURA.

\paragraph{AngularGrad}
AngularGrad \cite{Roy2021} rescales the Adam direction by an angle-like measure of consecutive gradients, the same kind of quantity that drives AURA.
The multiplier is $\phi^{\mathrm{Ang}}_{j,t} = \tfrac12+\tfrac12\tanh\bigl|f(A_{j,\min})\bigr|$, where $A_{j,t} = \arctan\bigl|(g_{j,t}-g_{j,t-1})/(1+g_{j,t}g_{j,t-1})\bigr|$ is the angle between two consecutive scalar gradients interpreted as slopes, $A_{j,\min} = \min\{A_{j,t-1},A_{j,t}\}$, and $f$ is either the cosine or the tangent.
Three differences from AURA stand out.
First, AngularGrad is designed and validated on real-valued networks only.
Second, $A_{j,t}$ is an unoriented slope angle, rather than the oriented phase difference of complex updates.
Third, $A_{j,t}$ hardly separates concordant from opposite gradients, because it compares slopes and a sign change of a small gradient is a small angle: with $f = \cos$, the pair $g_{j,t-1} = -0.01$, $g_{j,t} = 0.01$ gives $A_{j,t} = 0.02$, two equal gradients give $A_{j,t} = 0$, and both lead to $\phi^{\mathrm{Ang}}_{j,t} \approx 0.881$.

\paragraph{Hypergradient descent}
Hypergradient descent \cite{Gunes2018} differentiates the loss with respect to the step size.
Although conceptually distant from AURA, its multiplicative variant shares an important ingredient. As reported by Baydin et al.\ \cite{Gunes2018}, the multiplicative rule is
\begin{equation}
	\alpha_t = \alpha_{t-1}\left(1 - \beta' \cos\phi_t\right),
\end{equation}
where
\begin{equation}
	\cos\phi_t =
	\frac{
		\widetilde{\nabla}\loss(w_{t-1})^{\top}
		\nabla_{\alpha}u(\mathcal W_{t-2},\alpha_{t-1})
	}{
		\left\|\widetilde{\nabla}\loss(w_{t-1})\right\|
		\left\|\nabla_{\alpha}u(\mathcal W_{t-2},\alpha_{t-1})\right\|
	}.
\end{equation}
Here $\beta' \in \RR$ is a hyperparameter, $\loss \colon \RR^N \to \RR$ is the loss, $\widetilde{\nabla}\loss$ its stochastic estimator, and $u$ the update rule of the base method, $w_t = u(\mathcal W_{t-1},\alpha) \in \RR^N$ with $\mathcal W_t = \{w_i\}_{i=0}^{t}$.
The step size is thus updated according to the angle between two vectors, as in AURA; the vectors, however, differ in meaning.
Since $\nabla_\alpha u(\mathcal W_{t-2},\alpha_{t-1})$ is the previous update direction, hypergradient descent correlates the current gradient with the previous update, whereas AURA correlates two consecutive update directions.

\paragraph{Phase of the complex step size}
FGACGD \cite{Yin2026} also relies on an angle, but of a different kind: the phase $\phi_t \in [-\pi/4,\pi/4]$ of a single complex step size $|\eta_t|e^{\mathrm i\phi_t}$ shared by all trainable parameters.
The phase is selected by comparing the gradient norms at three trial points, reached with step sizes $\rho e^{\mathrm i0}$ and $\rho e^{\pm\mathrm i\pi/4}$, and steers the update toward flatter regions; no angle between consecutive directions is measured.

\paragraph{Summary}
The methods above measure the agreement of successive gradients or updates in two ways.
Sign-based rules \cite{Jacobs1988, Riedmiller1993, Eckhoff2023} detect only whether the direction is reversed, and angle-based rules \cite{Roy2021, Gunes2018, Malviya2024} detect an unoriented angle computed from real-valued gradients.
AURA operates per parameter in the complex plane and measures how much consecutive directions differ -- in length and in angle -- and in which sense they rotate.
This combination is the novelty claimed here; its ingredients are established: per-parameter multiplicative step control \cite{Riedmiller1993}, angle-sensitive modulation \cite{Roy2021, Gunes2018}, and the combination of Adam with RPROP \cite{Eckhoff2023}.

\section{Case studies}\label{sec:benchmarks}

This section compares AURA with the baseline optimizers on four test cases, selected to span different common learning tasks, architectures, and training regimes rather than to reach the best possible performance on any of them.
The first three approximate a non-holomorphic (\Cref{sec:test_case_non_holomorphic}), a holomorphic (\Cref{sec:test_case_holomorphic}), and a multivariate (\Cref{sec:test_case_multivariate}) complex function with fully connected networks; they are inexpensive and serve for sensitivity studies on the step size, the moment decay rates, the mini-batch size, and the floating-point precision.
The last one tests whether the conclusions carry over to a physics-informed loss with full-batch training (\Cref{sec:test_case_pinn}).
Since the object of study is the optimizer and not the task, reference architectures and standard setups are used throughout.

The protocol follows, as far as practicable, the recommendations of \cite{Schneider2018, Schmidt2021, McGreivy2024} on the fair comparison of optimizers.
Within each case, all methods start from the same initialization, are trained on the same data, and are run for several random seeds; they are compared through the training and test loss curves and the three scalar metrics of \Cref{subsec:metrics}: the best training loss reached, the area under the logarithmic learning curve, and the wall-clock time per step relative to plain stochastic gradient descent.
Since the hyperparameter search space can by itself determine the ranking of optimizers \cite{Choi2020}, the hyperparameters of AURA are fixed, although their number would allow per-case tuning; this choice penalizes AURA, if anything, and tests at the same time whether the method requires such tuning.
\Cref{sec:test_case_pinn} complements this with a comparison in which every optimizer is tuned under limited budget.

Four cases cannot cover every training scenario.
The aim is to show that AURA is advantageous on a reasonably broad set of problems under a fair protocol, to quantify its overhead, and to document a failure mode (\Cref{sec:test_case_holomorphic}); extending the comparison to further tasks and larger models is left to future work.

All the test cases were run on a single A100 40GB graphics processing unit (GPU).

\subsection{Metrics for comparison}\label{subsec:metrics}

In each test case, every method $m$ of the compared set $\mathcal M$ is trained from the same initialization and on the same data, independently for $N_{\rm seed}$ random seeds $s = 1,\ldots,N_{\rm seed}$; the value of $N_{\rm seed}$ is given in the settings table of each case.
Each run records the training loss $\loss_{\rm train}^{(m,s)}(t)$ at every optimizer step $t$, namely the loss of the current mini-batch, or of the whole training set in the full-batch case, and the test-set loss $\loss_{\rm test}^{(m,s)}(t)$ at regular intervals.
Methods are compared through the loss curves and through three scalar metrics, both summarized over seeds by the median and the 25th and 75th percentiles $Q_{25}$ and $Q_{75}$, computed by linear interpolation; the tables report a metric with median $x$ as $x^{+(Q_{75}-x)}_{-(x-Q_{25})}$.
Steps at which the loss is not finite, which occur only in diverged runs, are excluded from the scalar metrics but highlighted in the results summary tables.
The three metrics are listed below.
\begin{itemize}
	\item \textbf{Min training loss.} The best training loss reached within a run, $\loss_{\min}^{(m,s)} = \min_t\loss_{\rm train}^{(m,s)}(t)$, summarized over seeds by its median and percentiles,
	\begin{equation}\label{eq:benchmark_min_mean_loss}
		\loss_{\min}^{(m)} = \operatorname*{median}_{s = 1,\ldots,N_{\rm seed}}\loss_{\min}^{(m,s)}.
	\end{equation}
	It summarizes a run by a single number that does not depend on the choice of a stopping step. It is used to assess how effectively the optimizer minimizes the loss; the median reduces the influence of unusually favorable or unfavorable runs.
	\item \textbf{Area under the log learning curve.} To account for the whole trajectory rather than for its best point only, the training-loss curve is averaged on a base-10 logarithmic scale,
	\begin{equation}\label{eq:benchmark_naulc}
		A^{(m,s)} = \frac{1}{T+1}\sum_{t = 0}^{T}\Bigl(C + \log_{10}\loss_{\rm train}^{(m,s)}(t)\Bigr), \qquad C = 10,
	\end{equation}
	where $T$ is the final optimizer step, common to all methods within a case.
	$A^{(m,s)}$ is the normalized area between the log learning curve and a floor at loss $10^{-C}$, placed below the smallest loss reached in any case so that the area is non-negative; equivalently, it is $C$ plus the base-10 logarithm of the geometric-mean training loss, and the constant $C$ affects no comparison.
	Since every decade of loss reduction has the same weight, $A^{(m,s)}$ rewards methods that reduce the loss early and keep it low, and is not dominated by the large losses of the initial transient. This metric is used as it provides a quantitative assessment of the whole training curve. A complementary qualitative assessment is provided by the training-curve plots.
	\item \textbf{Training time.} The wall-clock time $\tau^{(m,s)}$ of the training loop, excluding the diagnostic instrumentation, such as the routines that store the data from which the figures are generated, and the just-in-time compilation of the first step.	Since a time in seconds is tied to the GPU it was measured on, following \cite{Schneider2018}, we use the ratio
	\begin{equation}\label{eq:benchmark_training_time_normalized}
		\tilde\tau^{(m,s)} = \tau^{(m,s)} \big/ \tau^{(\mathrm{SGD})},
	\end{equation}
	summarized over seeds by its median $\tilde\tau^{(m)}$ and percentiles, where $\tau^{(\mathrm{SGD})}$ is the median time of the plain stochastic gradient descent update, $w_{j,t+1} = w_{j,t} - \alpha\, g_{j,t}$, over $N_{\mathrm{SGD}} = 5$ seeds, run with the same architecture, precision, step budget, and GPU.
	As SGD is the cheapest first-order step, $\tilde\tau^{(m)}$ is the (almost) hardware-independent factor by which method $m$ is slower than it.
	A separate SGD baseline is measured for each setting whose per-step cost differs, i.e., for each architecture and for \texttt{complex64} versus \texttt{complex128}.
	SGD serves only as a unit of time and is not part of $\mathcal M$. We adopt this metric because the actual training time is an important practical feature of the optimizer reflecting its computational efficiency. 
\end{itemize}

\subsection{Optimizer settings}\label{subsec:optimizer_settings}
The baseline optimizers are chosen as follows.
RPROP \cite{Riedmiller1993} is included because, as in AURA, the agreement of successive derivatives sets the size of the next step. 
Adam \cite{Kingma2014} and NadamW, that is, Nadam \cite{Dozat2016} with decoupled weight decay \cite{Loshchilov2018}, are included because they are widely used and perform well for training complex-valued networks \cite{Mayer2025}; Adam is, in addition, the base optimizer of Adam-AURA, so that the comparison between the two isolates the effect of AURA.
CvAMSGrad, the split-complex extension of AMSGrad \cite{Sashank2018} proposed in \cite{Mayer2025}, is included because it attains a lower steady-state error than SGD on the two complex-valued benchmark problems of \cite{Mayer2025}, at a moderate increase in computational cost.
Muon \cite{Jordan2024} is included because it is designed specifically for the weight matrices of the hidden layers of neural networks, and it is the base optimizer of Muon-AURA. 

The default hyperparameters for each optimizer are listed in \Cref{tab:optimizer_settings_shared}. Where applicable, these are adjusted as described in the relevant test case.
The ten hyperparameters specific to AURA are held fixed across all test cases, except in \Cref{sec:test_case_pinn}, where its principal ones are tuned together with those of the baselines. The resulting comparison is pragmatic and fair: if anything, this constraint disadvantages AURA, which nonetheless outperforms the baselines in many cases.

\begin{table*}[!t]
\centering
\caption{Default hyperparameters used in the benchmark cases in \Cref{sec:benchmarks}.}
\label{tab:optimizer_settings_shared}
\small
\begin{tabularx}{\linewidth}{@{}lX@{}}
\toprule
Optimizer & Hyperparameters \\
\midrule
RPROP & $\eta_-=0.5$, $\eta_+=1.2$, $\Delta\in[1\times 10^{-6},50]$; $\Delta_0=\alpha$ \\
Adam & $\beta_1=0.9$, $\beta_2=0.999$, $\eps_{\mathrm A}=1\times 10^{-8}$ \\
Adam (variable LR) & Same $\beta_1,\beta_2,\eps_{\mathrm A}$ as Adam; learning rate $\alpha$ for the first $50\%$ of updates, then $\alpha/10$ \\
NadamW & $\beta_1=0.9$, $\beta_2=0.999$, $\eps_{\mathrm A}=1\times 10^{-8}$, weight decay: $1\times 10^{-4}$ \\
CvAMSGrad & $\beta_1=0.9$, $\beta_2=0.999$, $\eps_{\mathrm A}=1\times 10^{-8}$; bias correction included \\
Muon & $\beta=0.95$, Newton--Schulz steps: 5, Nesterov: yes, weight decay: $0$; matrix step $10\,\alpha$; bias fallback AdamW at $\beta_1=0.9$, $\beta_2=0.999$, $\eps_{\mathrm A}=1\times 10^{-8}$ \\
Adam-AURA & $\beta_\zeta=0.95$, $\eps_{\mathrm E}=1\times 10^{-6}$, $\chi_\mathrm{a}=0.7$, $\chi_\mathrm{o}=0.4$, $\Psi_\mathrm{a}=0.015$, $\Psi_\mathrm{o}=0.3$, $\eta_-=0.99$, $\eta_+=1.01$, $\gamma\in[1\times 10^{-3},1000]$, $\lambda=1\times 10^{-4}$ \\
Muon-AURA & $\beta=0.95$, Newton--Schulz steps: 5, Nesterov: yes, matrix step $10\,\alpha$; $\beta_\zeta=0.95$, $\eps_{\mathrm E}=1\times 10^{-6}$, $\chi_\mathrm{a}=0.75$, $\chi_\mathrm{o}=0.4$, $\Psi_\mathrm{a}=0.01$, $\Psi_\mathrm{o}=0.2$, $\eta_-=0.99$, $\eta_+=1.01$, $\gamma\in[1\times 10^{-3},1000]$, $\lambda=1\times 10^{-4}$; bias direction Adam at $\beta_1=0.9$, $\beta_2=0.999$, $\eps_{\mathrm A}=1\times 10^{-8}$ \\
\bottomrule
\end{tabularx}
\end{table*}

\subsection{TEST 1: univariate scalar complex non-holomorphic function}\label{sec:test_case_non_holomorphic}

\paragraph{Rationale}
This test provides a lightweight analytical benchmark for sensitivity and ablation studies. The goal is to approximate a non-holomorphic function, complementing the holomorphic case considered in a separate test. We empirically investigate the sensitivity of the optimizers to their main hyperparameters by comparing their performance across two fully connected neural network architectures, different step sizes, and different exponential moving average (EMA) decay rates.

\paragraph{Learning task}
The domain is the square
$\Omega_\zeta = \{\zeta\in\CC:|\real\zeta| \leq 1,\ |\imag\zeta| \leq 1\}$.
The target function is
\begin{align}
	f(\zeta) ={}&
	\exp\!\big((0.30-0.20\mathrm i)\,\zeta\overline\zeta\big)
	+0.25\sin(\zeta)\cos(\overline\zeta) \notag\\
	&+0.10\,\zeta^2\overline\zeta
	+0.08\,\zeta\overline\zeta^2
	-0.05\mathrm i\,(\zeta\overline\zeta)^2.
	\label{eq:optimizer_non_holomorphic_coupled_target}
\end{align}

The training and test sets consist of mutually independent samples drawn uniformly from $\Omega_\zeta$:
\begin{align}
	\mathcal Z_{\rm train} &= \{\zeta_i\}_{i=1}^{n_{\rm train}}, &
	\zeta_i &\overset{\text{i.i.d.}}{\sim}\operatorname{Unif}(\Omega_\zeta),
	\label{eq:benchmark_train_set}\\
	\mathcal Z_{\rm test} &= \{\zeta_i'\}_{i=1}^{n_{\rm test}}, &
	\zeta_i' &\overset{\text{i.i.d.}}{\sim}\operatorname{Unif}(\Omega_\zeta),
	\qquad \mathcal Z_{\rm test} \cap \mathcal Z_{\rm train} = \emptyset,
	\label{eq:benchmark_test_set}
\end{align}
with $n_{\rm test} = \operatorname{round}(0.2\,n_{\rm train})$.
The real and imaginary parts of each sample are drawn independently from
$\operatorname{Unif}([-1,1])$.
Denoting the neural network by $f_{\mathcal W}$ and a nonempty mini-batch by
$\mathcal B\subseteq\mathcal Z_{\rm train}$, training minimizes the mean squared error
\begin{equation}
	\loss_{\mathcal B}(\mathcal W)
	= \frac{1}{|\mathcal B|}
	\sum_{\zeta\in\mathcal B}
	\bigl|f_{\mathcal W}(\zeta)-f(\zeta)\bigr|^2.
	\label{eq:benchmark_batch_loss}
\end{equation}

\paragraph{Network architectures}
The networks are compositions of affine maps and componentwise nonlinear activations:
\begin{equation*}
	f_{\mathcal W}
	= A_L\circ\varsigma\circ A_{L-1}\circ\cdots
	\circ\varsigma\circ A_1,
\end{equation*}
with $A_\ell(z)=W_\ell z+b_\ell$, $\ell=1,\ldots,L$,
where $W_\ell\in\CC^{n_\ell\times n_{\ell-1}}$,
$b_\ell\in\CC^{n_\ell}$, $n_0=n_L=1$, and
$\mathcal W=\{W_\ell,b_\ell\}_{\ell=1}^L$.
The activation
\begin{equation*}
	\varsigma(\zeta)
	= \operatorname{SiLU}(\real\zeta)
	+\mathrm i\,\operatorname{SiLU}(\imag\zeta),
\end{equation*}
is applied componentwise in the hidden layers. This activation is non-holomorphic, so the networks are generally non-holomorphic.

The complete benchmark configuration is given in
\Cref{tab:optimizer_settings_non_holomorphic}.

\begin{table}[!ht]
\centering
\caption{Configuration of the Non-holomorphic optimizer benchmark.}
\label{tab:optimizer_settings_non_holomorphic}
\small
\begin{tabularx}{\linewidth}{@{}>{\raggedright\arraybackslash}p{0.32\linewidth}X@{}}
\toprule
Setting & Value \\
\midrule
Primary architecture & $(1,32,32,32,32,1)$, \numprint{3265} trainable parameters \\
Secondary architecture & $(1,128,128,128,128,128,1)$, \numprint{66433} trainable parameters \\
Hidden activation & $\operatorname{SiLU}(\real\zeta)+\mathrm i\operatorname{SiLU}(\imag\zeta)$ \\
Training / test set & \numprint{2500} / $500$ points (test: 20\%) \\
Mini-batch size & $256$ points \\
Updates / seeds & \numprint{12000} / $\{0,\ldots,4\}$ \\
Arithmetic & \texttt{complex64} (32-bit real components) \\
Initialization & Glorot/Xavier normal rule \cite{Glorot2010}, applied separately to the real and imaginary parts; zero biases \\
Base learning rate $\alpha$ (primary) & $5\times 10^{-4}$ \\
Secondary-architecture learning rate & $5\times 10^{-5}$ \\
Optimizer hyperparameters & default values; see \Cref{tab:optimizer_settings_shared} \\
\bottomrule
\end{tabularx}
\end{table}

\FloatBarrier

\begin{figure*}[!t]
	\centering
	\textbf{Loss: TEST 1 -- non-holomorphic}\par\vspace{0.3\baselineskip}
	\begin{minipage}[c]{0.02\linewidth}
		\centering
		\rotatebox{90}{Primary architecture}
	\end{minipage}%
	\begin{minipage}[c]{0.98\linewidth}
		\centering
		\begin{minipage}[c]{0.02\linewidth}
			\centering
			\rotatebox{90}{$\alpha/10$}
		\end{minipage}%
		\begin{minipage}[c]{0.98\linewidth}
			\centering
			\includegraphics[width=0.80\linewidth,page=1]{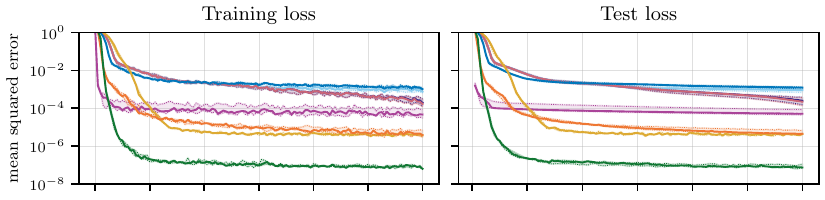}
		\end{minipage}\\
		\begin{minipage}[c]{0.02\linewidth}
			\centering
			\rotatebox{90}{$\alpha$}
		\end{minipage}%
		\begin{minipage}[c]{0.98\linewidth}
			\centering
			\includegraphics[width=0.80\linewidth,page=2]{figures/optimizer_losses_non_holomorphic.pdf}
		\end{minipage}\\
		\begin{minipage}[c]{0.02\linewidth}
			\centering
			\rotatebox{90}{$10\alpha$}
		\end{minipage}%
		\begin{minipage}[c]{0.98\linewidth}
			\centering
			\includegraphics[width=0.80\linewidth,page=3]{figures/optimizer_losses_non_holomorphic.pdf}
		\end{minipage}
	\end{minipage}\\[0.8\baselineskip]
	\begin{minipage}[c]{0.02\linewidth}
		\centering
		\rotatebox{90}{Secondary architecture}
	\end{minipage}%
	\begin{minipage}[c]{0.98\linewidth}
		\centering
		\begin{minipage}[c]{0.02\linewidth}
			\centering
			\rotatebox{90}{$\alpha/10$}
		\end{minipage}%
		\begin{minipage}[c]{0.98\linewidth}
			\centering
			\includegraphics[width=0.80\linewidth,page=4]{figures/optimizer_losses_non_holomorphic.pdf}
		\end{minipage}\\
		\begin{minipage}[c]{0.02\linewidth}
			\centering
			\rotatebox{90}{$\alpha$}
		\end{minipage}%
		\begin{minipage}[c]{0.98\linewidth}
			\centering
			\includegraphics[width=0.80\linewidth,page=5]{figures/optimizer_losses_non_holomorphic.pdf}
		\end{minipage}\\
		\begin{minipage}[c]{0.02\linewidth}
			\centering
			\rotatebox{90}{$10\alpha$}
		\end{minipage}%
		\begin{minipage}[c]{0.98\linewidth}
			\centering
			\includegraphics[width=0.80\linewidth,page=6]{figures/optimizer_losses_non_holomorphic.pdf}
		\end{minipage}
	\end{minipage}
	\par\vspace{0.4\baselineskip}
	\includegraphics{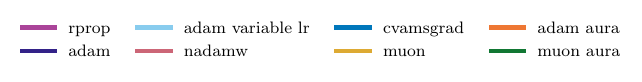}
	\par\vspace{0.2\baselineskip}
	\caption{Training-set and test-set mean squared errors for the
	Non-holomorphic optimizer benchmark. The two architecture blocks show
	the network trained at the primary (top) and secondary (bottom)
	architecture (\Cref{tab:optimizer_settings_non_holomorphic}).
	Within each block, the rows sweep the base learning rate $\alpha$ across
	$\alpha/10$, $\alpha$ and $10\alpha$ from top to bottom. Solid curves
	show the pointwise median over all
	$5$ seeds; shaded regions span the interquartile range (25th--75th	
	percentile) over seeds, whose edges are also drawn as dotted curves.}
	\label{fig:optimizer_losses_non_holomorphic}
\end{figure*}

\paragraph{Results -- $\alpha$ sensitivity}
\Cref{tab:optimizer_training_time_non_holomorphic} reports the three metrics for the step sizes $\alpha/10$, $\alpha$, and $10\alpha$, and \Cref{fig:optimizer_losses_non_holomorphic} shows the corresponding training- and test-loss curves for both architectures.
Over this $100$-fold range, $\loss_{\min}^{(m)}$ varies by more than two orders of magnitude for Adam and its variants and by more than three for Muon at the primary architecture and by more than one at the secondary, but by less than one for Adam-AURA, and it remains of order $10^{-7}$ or below for Muon-AURA.
RPROP is similarly insensitive to $\alpha$, but its minimum loss is $6$ to $50$ times that of Adam-AURA.
Muon-AURA attains the lowest $\loss_{\min}^{(m)}$ and $A^{(m)}$ in all six settings; since $A^{(m)}$ is $C$ plus the base-10 logarithm of the geometric-mean training loss \eqref{eq:benchmark_naulc}, its margin of about $2$ to $3$ over the best baseline corresponds to a loss roughly two to three orders of magnitude lower along the whole trajectory, not only at its best point.
This improvement comes at a small cost: the time ratio \eqref{eq:benchmark_training_time_normalized} of Adam-AURA exceeds that of Adam by at most $0.1$, within the spread over seeds, and that of Muon-AURA is even slightly below that of Muon, so the higher ratio of Muon-AURA, up to $2.1$, stems from the Muon direction rather than from AURA.

\begin{table*}[!t]
\centering
\caption{Step-size sensitivity for the Non-holomorphic benchmark.}
\label{tab:optimizer_training_time_non_holomorphic}
\footnotesize
\setlength{\tabcolsep}{3pt}
\begin{adjustbox}{max width=\linewidth,center}
\begin{tabular}{@{}ll*{7}{c}@{}}
\toprule
 & & \multicolumn{3}{c}{$\mathscr{L}_{\min}^{(m)}\ (\times10^{-6})$} & \multicolumn{3}{c}{$A^{(m)}$} & Training time ($\times$SGD) \\
\cmidrule(lr){3-5}\cmidrule(lr){6-8}\cmidrule(lr){9-9}
Arch. & Optimizer & $\alpha/10$ & $\alpha$ & $10\alpha$ & $\alpha/10$ & $\alpha$ & $10\alpha$ & $\alpha$ \\
\midrule
\multirow{8}{*}{Primary} & RPROP & \mbox{$34^{+19}_{-8}$} & \mbox{$203^{+6}_{-93}$} & \mbox{$41^{+6}_{-14}$} & \mbox{$5.9^{+0.3}_{-0.1}$} & \mbox{$6.6^{+0.01}_{-0.2}$} & \mbox{$6.1^{+0.1}_{-0.3}$} & \mbox{$1.67^{+0.0002}_{-0.02}$} \\
 & Adam & \mbox{$167^{+92}_{-52}$} & \mbox{$4.6^{+0.5}_{-0.1}$} & \mbox{$1.3^{+0.1}_{-0.2}$} & \mbox{$7.3^{+0.1}_{-0.1}$} & \mbox{$5.8^{+0.1}_{-0.03}$} & \mbox{$5.4^{+0.01}_{-0.04}$} & \mbox{$1.59^{+0.005}_{-0.04}$} \\
 & Adam (variable LR) & \mbox{$614^{+106}_{-116}$} & \mbox{$7.6^{+2.5}_{-0.4}$} & \mbox{$1.5^{+0.2}_{-0.03}$} & \mbox{$7.4^{+0.1}_{-0.1}$} & \mbox{$5.8^{+0.1}_{-0.01}$} & \mbox{$5.1^{+0.002}_{-0.001}$} & \mbox{$\bm{1.580^{+0.006}_{-0.012}}$} \\
 & NadamW & \mbox{$151^{+99}_{-49}$} & \mbox{$5.6^{+0.7}_{-0.2}$} & \mbox{$1.1^{+0.1}_{-0.02}$} & \mbox{$7.3^{+0.1}_{-0.1}$} & \mbox{$5.9^{+0.1}_{-0.05}$} & \mbox{$5.8^{+0.02}_{-0.03}$} & \mbox{$1.59^{+0.04}_{-0.01}$} \\
 & CvAMSGrad & \mbox{$893^{+119}_{-94}$} & \mbox{$16^{+0.3}_{-4}$} & \mbox{$2.2^{+0.3}_{-0.02}$} & \mbox{$7.4^{+0.05}_{-0.002}$} & \mbox{$6.1^{+0.05}_{-0.1}$} & \mbox{$5.1^{+0.01}_{-0.04}$} & \mbox{$1.6^{+0.1}_{-0.2}$} \\
 & Muon & \mbox{$0.2^{+0.4}_{-0.01}$} & \mbox{$32^{+3}_{-0.4}$} & \mbox{$283^{+60}_{-44}$} & \mbox{$5.3^{+0.02}_{-0.003}$} & \mbox{$6.0^{+0.01}_{-0.002}$} & \mbox{$7.0^{+0.001}_{-0.01}$} & \mbox{$1.88^{+0.0002}_{-0.02}$} \\
 & Adam-AURA & \mbox{$3.0^{+2.1}_{-0.4}$} & \mbox{$3.9^{+0.1}_{-1.9}$} & \mbox{$1.7^{+0.8}_{-0.04}$} & \mbox{$5.3^{+0.1}_{-0.1}$} & \mbox{$5.1^{+0.1}_{-0.1}$} & \mbox{$4.8^{+0.03}_{-0.1}$} & \mbox{$1.64^{+0.11}_{-0.01}$} \\
 & Muon-AURA & \mbox{$\bm{0.1^{+0.003}_{-0.002}}$} & \mbox{$\bm{0.04^{+0.004}_{-0.005}}$} & \mbox{$\bm{0.01^{+0.001}_{-0.0001}}$} & \mbox{$\bm{3.4^{+0.1}_{-0.003}}$} & \mbox{$\bm{3.1^{+0.04}_{-0.1}}$} & \mbox{$\bm{2.7^{+0.003}_{-0.01}}$} & \mbox{$1.80^{+0.09}_{-0.03}$} \\
\cmidrule(lr){1-9}
\multirow{8}{*}{Secondary} & RPROP & \mbox{$9.2^{+19.1}_{-0.2}$} & \mbox{$16^{+4}_{-2}$} & \mbox{$17^{+1}_{-1}$} & \mbox{$5.4^{+0.4}_{-0.04}$} & \mbox{$5.7^{+0.03}_{-0.1}$} & \mbox{$5.6^{+0.1}_{-0.05}$} & \mbox{$1.70^{+0.02}_{-0.01}$} \\
 & Adam & \mbox{$1505^{+4}_{-103}$} & \mbox{$28^{+4}_{-8}$} & \mbox{$3.4^{+0.2}_{-0.1}$} & \mbox{$8.0^{+0.01}_{-0.04}$} & \mbox{$6.8^{+0.02}_{-0.05}$} & \mbox{$5.9^{+0.01}_{-0.1}$} & \mbox{$1.65^{+0.08}_{-0.01}$} \\
 & Adam (variable LR) & \mbox{$1786^{+21}_{-6}$} & \mbox{$305^{+12}_{-23}$} & \mbox{$4.4^{+0.3}_{-0.6}$} & \mbox{$8.0^{+0.01}_{-0.04}$} & \mbox{$7.0^{+0.02}_{-0.02}$} & \mbox{$5.7^{+0.01}_{-0.2}$} & \mbox{$\bm{1.64^{+0.01}_{-0.02}}$} \\
 & NadamW & \mbox{$1494^{+9}_{-93}$} & \mbox{$27^{+1}_{-5}$} & \mbox{$3.7^{+0.8}_{-0.01}$} & \mbox{$8.0^{+0.01}_{-0.04}$} & \mbox{$6.8^{+0.02}_{-0.03}$} & \mbox{$6.1^{+0.01}_{-0.03}$} & \mbox{$1.64^{+0.01}_{-0.02}$} \\
 & CvAMSGrad & \mbox{$1785^{+16}_{-24}$} & \mbox{$287^{+18}_{-14}$} & \mbox{$4.1^{+0.2}_{-0.7}$} & \mbox{$7.9^{+0.01}_{-0.05}$} & \mbox{$7.0^{+0.04}_{-0.01}$} & \mbox{$5.7^{+0.03}_{-0.1}$} & \mbox{$1.66^{+0.002}_{-0.02}$} \\
 & Muon & \mbox{$50^{+6}_{-14}$} & \mbox{$3.5^{+0.3}_{-0.3}$} & \mbox{$55^{+0.1}_{-0.3}$} & \mbox{$8.4^{+0.01}_{-0.1}$} & \mbox{$5.2^{+0.02}_{-0.01}$} & \mbox{$6.1^{+0.01}_{-0.01}$} & \mbox{$2.218^{+0.010}_{-0.0003}$} \\
 & Adam-AURA & \mbox{$1.5^{+0.1}_{-0.2}$} & \mbox{$1.6^{+0.2}_{-0.4}$} & \mbox{$0.6^{+0.2}_{-0.2}$} & \mbox{$5.4^{+0.01}_{-0.1}$} & \mbox{$4.9^{+0.2}_{-0.01}$} & \mbox{$4.6^{+0.1}_{-0.01}$} & \mbox{$1.68^{+0.05}_{-0.05}$} \\
 & Muon-AURA & \mbox{$\bm{0.02^{+0.001}_{-0.001}}$} & \mbox{$\bm{0.01^{+0.0004}_{-0.0003}}$} & \mbox{$\bm{0.01^{+0.0004}_{-0.000004}}$} & \mbox{$\bm{2.9^{+0.03}_{-0.02}}$} & \mbox{$\bm{2.8^{+0.02}_{-0.01}}$} & \mbox{$\bm{2.6^{+0.01}_{-0.04}}$} & \mbox{$2.147^{+0.015}_{-0.003}$} \\
\bottomrule
\end{tabular}
\end{adjustbox}
\end{table*}

\FloatBarrier

\FloatBarrier

\begin{figure*}[!t]
	\centering
	\textbf{Loss: TEST 1 -- non-holomorphic ($\beta_1$, $\beta_2$ sweep)}\par\vspace{0.3\baselineskip}
	\begin{minipage}[c]{0.02\linewidth}
		\centering
		\rotatebox{90}{Primary architecture}
	\end{minipage}%
	\begin{minipage}[c]{0.98\linewidth}
		\centering
		\subfloat[$\beta_1=0.85,\ \beta_2=0.99$]{\includegraphics[width=0.32\linewidth,page=1]{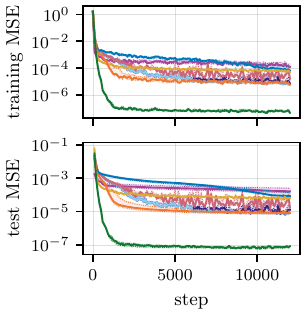}}\hfill
		\subfloat[$\beta_1=0.9,\ \beta_2=0.99$]{\includegraphics[width=0.32\linewidth,page=4]{figures/optimizer_beta_sweep_non_holomorphic.pdf}}\hfill
		\subfloat[$\beta_1=0.95,\ \beta_2=0.99$]{\includegraphics[width=0.32\linewidth,page=7]{figures/optimizer_beta_sweep_non_holomorphic.pdf}}\\
		\subfloat[$\beta_1=0.85,\ \beta_2=0.999$]{\includegraphics[width=0.32\linewidth,page=2]{figures/optimizer_beta_sweep_non_holomorphic.pdf}}\hfill
		\subfloat[$\beta_1=0.9,\ \beta_2=0.999$]{\includegraphics[width=0.32\linewidth,page=5]{figures/optimizer_beta_sweep_non_holomorphic.pdf}}\hfill
		\subfloat[$\beta_1=0.95,\ \beta_2=0.999$]{\includegraphics[width=0.32\linewidth,page=8]{figures/optimizer_beta_sweep_non_holomorphic.pdf}}\\
		\subfloat[$\beta_1=0.85,\ \beta_2=0.9999$]{\includegraphics[width=0.32\linewidth,page=3]{figures/optimizer_beta_sweep_non_holomorphic.pdf}}\hfill
		\subfloat[$\beta_1=0.9,\ \beta_2=0.9999$]{\includegraphics[width=0.32\linewidth,page=6]{figures/optimizer_beta_sweep_non_holomorphic.pdf}}\hfill
		\subfloat[$\beta_1=0.95,\ \beta_2=0.9999$]{\includegraphics[width=0.32\linewidth,page=9]{figures/optimizer_beta_sweep_non_holomorphic.pdf}}
	\end{minipage}
	\par\vspace{0.4\baselineskip}
	\includegraphics{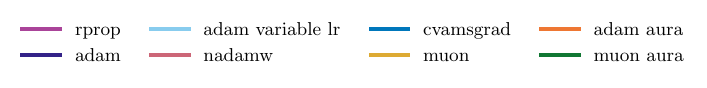}
	\par\vspace{0.2\baselineskip}
	\caption{Training-set and test-set mean squared errors for the
	Non-holomorphic benchmark over the $3\times3$ $\beta_1$/$\beta_2$
	grid. Rows fix $\beta_2$ ($0.99$, $0.999$, $0.9999$);
	columns fix $\beta_1$ ($0.85$, $0.9$, $0.95$). The step size is fixed to its baseline, and each panel shows the training error above the test error. Solid curves show the
	pointwise median over the $3$ seeds; shaded regions span the
	interquartile range (25th--75th percentile) over seeds.}
	\label{fig:optimizer_beta_sweep_non_holomorphic}
\end{figure*}

\paragraph{Results -- $\beta_1$ and $\beta_2$ sensitivity}
\Cref{tab:optimizer_beta_sweep_non_holomorphic} repeats the comparison over a $3\times3$ grid of the first- and second-moment exponential-decay rates $\beta_1\in\{0.85,0.9,0.95\}$ and $\beta_2\in\{0.99,0.999,0.9999\}$, at the base step size $\alpha$ and the primary architecture, with $3$ seeds per pair, and \Cref{fig:optimizer_beta_sweep_non_holomorphic} shows the corresponding training- and test-loss curves.
The two rates enter Adam, Adam (variable LR), NadamW, CvAMSGrad, and the direction of Adam-AURA, whereas they reach Muon and Muon-AURA only through the Adam-type update of the biases, and RPROP does not use them.
The RPROP column nonetheless differs between pairs, for example $110$ and $128$ in $\loss_{\min}^{(m)}$, because each pair is a separate run and we did not set the XLA compiler used by JAX \cite{jax2018github} to be fully deterministic on the GPU.
Muon-AURA attains the lowest $\loss_{\min}^{(m)}$ and $A^{(m)}$ at every pair and is insensitive to both rates, with $A^{(m)}$ between $3.0$ and $3.1$, whereas the $\loss_{\min}^{(m)}$ of Muon, exposed to them in the same way, varies by more than a factor of five.
Adam-AURA has the second lowest $A^{(m)}$ at every pair and benefits most from a larger $\beta_1$: raising $\beta_1$ from $0.85$ to $0.95$ lowers its $\loss_{\min}^{(m)}$ by a factor of about $8$ to $12$ at each $\beta_2$, against at most $1.7$ for Adam.
This is consistent with the mechanism of AURA, since a longer averaging window makes consecutive Adam directions agree more often, which increases the multiplier; the default value $\beta_1=0.9$ of \Cref{tab:optimizer_settings_shared} therefore leaves this gain unexploited.

\begin{table*}[!t]
\centering
\caption{$\beta_1$/$\beta_2$ sensitivity for the Non-holomorphic benchmark.}
\label{tab:optimizer_beta_sweep_non_holomorphic}
\small
\setlength{\tabcolsep}{4pt}
\begin{adjustbox}{max width=\textwidth,center}
\begin{tabular}{@{}ll*{8}{c}@{}}
\toprule
$\beta_1$ & $\beta_2$ & \multicolumn{1}{c}{\rotatebox[origin=l]{90}{RPROP}} & \multicolumn{1}{c}{\rotatebox[origin=l]{90}{Adam}} & \multicolumn{1}{c}{\rotatebox[origin=l]{90}{Adam (variable LR)}} & \multicolumn{1}{c}{\rotatebox[origin=l]{90}{NadamW}} & \multicolumn{1}{c}{\rotatebox[origin=l]{90}{CvAMSGrad}} & \multicolumn{1}{c}{\rotatebox[origin=l]{90}{Muon}} & \multicolumn{1}{c}{\rotatebox[origin=l]{90}{Adam-AURA}} & \multicolumn{1}{c}{\rotatebox[origin=l]{90}{Muon-AURA}} \\
\midrule
 & & \multicolumn{8}{c}{$\mathscr{L}_{\min}^{(m)}\ (\times10^{-6})$} \\
\cmidrule(lr){3-10}
$0.85$ & $0.99$ & \mbox{$110^{+49}_{-15}$} & \mbox{$4.4^{+0.2}_{-0.4}$} & \mbox{$5.5^{+0.4}_{-0.7}$} & \mbox{$5.2^{+1.9}_{-0.2}$} & \mbox{$55^{+37}_{-5}$} & \mbox{$41^{+0.2}_{-4}$} & \mbox{$5.8^{+1.1}_{-1.0}$} & \mbox{$\bm{0.04^{+0.01}_{-0.001}}$} \\
$0.9$ & $0.99$ & \mbox{$110^{+50}_{-15}$} & \mbox{$3.8^{+0.04}_{-0.4}$} & \mbox{$4.9^{+0.03}_{-0.5}$} & \mbox{$5.1^{+0.4}_{-0.8}$} & \mbox{$58^{+45}_{-0.1}$} & \mbox{$34^{+2}_{-2}$} & \mbox{$2.0^{+0.6}_{-0.3}$} & \mbox{$\bm{0.04^{+0.002}_{-0.003}}$} \\
$0.95$ & $0.99$ & \mbox{$110^{+50}_{-15}$} & \mbox{$2.6^{+0.4}_{-0.03}$} & \mbox{$4.1^{+0.1}_{-0.4}$} & \mbox{$3.3^{+0.2}_{-0.4}$} & \mbox{$151^{+29}_{-37}$} & \mbox{$30^{+1}_{-0.1}$} & \mbox{$0.5^{+0.8}_{-0.04}$} & \mbox{$\bm{0.03^{+0.003}_{-0.002}}$} \\
$0.85$ & $0.999$ & \mbox{$110^{+50}_{-15}$} & \mbox{$5.7^{+0.2}_{-0.4}$} & \mbox{$8.0^{+1.4}_{-0.1}$} & \mbox{$7.5^{+2.2}_{-0.4}$} & \mbox{$15^{+2}_{-2}$} & \mbox{$39^{+1}_{-3}$} & \mbox{$5.3^{+2.2}_{-1.0}$} & \mbox{$\bm{0.04^{+0.01}_{-0.003}}$} \\
$0.9$ & $0.999$ & \mbox{$128^{+41}_{-24}$} & \mbox{$4.9^{+0.1}_{-0.2}$} & \mbox{$7.6^{+1.3}_{-0.2}$} & \mbox{$6.6^{+0.4}_{-0.6}$} & \mbox{$16^{+2}_{-2}$} & \mbox{$37^{+0.1}_{-3}$} & \mbox{$1.9^{+1.1}_{-0.1}$} & \mbox{$\bm{0.04^{+0.01}_{-0.001}}$} \\
$0.95$ & $0.999$ & \mbox{$110^{+50}_{-15}$} & \mbox{$4.1^{+0.1}_{-0.1}$} & \mbox{$9.3^{+0.02}_{-1.1}$} & \mbox{$4.4^{+0.3}_{-0.01}$} & \mbox{$21^{+2}_{-2}$} & \mbox{$27^{+2}_{-1}$} & \mbox{$0.6^{+0.8}_{-0.1}$} & \mbox{$\bm{0.03^{+0.003}_{-0.0002}}$} \\
$0.85$ & $0.9999$ & \mbox{$128^{+41}_{-24}$} & \mbox{$9.2^{+0.5}_{-0.3}$} & \mbox{$35^{+1}_{-5}$} & \mbox{$11^{+2}_{-0.5}$} & \mbox{$6.5^{+1.0}_{-0.3}$} & \mbox{$30^{+3}_{-7}$} & \mbox{$10^{+0.3}_{-2}$} & \mbox{$\bm{0.04^{+0.01}_{-0.003}}$} \\
$0.9$ & $0.9999$ & \mbox{$110^{+50}_{-15}$} & \mbox{$8.7^{+0.5}_{-0.4}$} & \mbox{$36^{+0.4}_{-5}$} & \mbox{$9.3^{+0.6}_{-0.5}$} & \mbox{$6.5^{+0.7}_{-0.2}$} & \mbox{$20^{+3}_{-7}$} & \mbox{$3.1^{+1.2}_{-0.1}$} & \mbox{$\bm{0.04^{+0.01}_{-0.001}}$} \\
$0.95$ & $0.9999$ & \mbox{$110^{+50}_{-15}$} & \mbox{$10^{+1}_{-0.1}$} & \mbox{$40^{+6}_{-2}$} & \mbox{$9.7^{+0.5}_{-0.3}$} & \mbox{$7.4^{+0.5}_{-0.4}$} & \mbox{$7.9^{+10.2}_{-0.9}$} & \mbox{$1.2^{+0.8}_{-0.1}$} & \mbox{$\bm{0.03^{+0.004}_{-0.001}}$} \\
\midrule
 & & \multicolumn{8}{c}{$A^{(m)}$} \\
\cmidrule(lr){3-10}
$0.85$ & $0.99$ & \mbox{$6.4^{+0.1}_{-0.1}$} & \mbox{$5.6^{+0.1}_{-0.05}$} & \mbox{$5.5^{+0.1}_{-0.1}$} & \mbox{$5.9^{+0.1}_{-0.04}$} & \mbox{$6.6^{+0.1}_{-0.1}$} & \mbox{$6.0^{+0.005}_{-0.01}$} & \mbox{$5.2^{+0.1}_{-0.1}$} & \mbox{$\bm{3.1^{+0.04}_{-0.03}}$} \\
$0.9$ & $0.99$ & \mbox{$6.4^{+0.1}_{-0.1}$} & \mbox{$5.5^{+0.04}_{-0.05}$} & \mbox{$5.4^{+0.04}_{-0.1}$} & \mbox{$5.7^{+0.1}_{-0.04}$} & \mbox{$6.7^{+0.05}_{-0.1}$} & \mbox{$6.0^{+0.01}_{-0.01}$} & \mbox{$4.9^{+0.1}_{-0.1}$} & \mbox{$\bm{3.1^{+0.003}_{-0.04}}$} \\
$0.95$ & $0.99$ & \mbox{$6.4^{+0.1}_{-0.1}$} & \mbox{$5.4^{+0.04}_{-0.1}$} & \mbox{$5.4^{+0.03}_{-0.1}$} & \mbox{$5.4^{+0.04}_{-0.05}$} & \mbox{$6.8^{+0.02}_{-0.1}$} & \mbox{$5.9^{+0.01}_{-0.01}$} & \mbox{$4.4^{+0.2}_{-0.03}$} & \mbox{$\bm{3.0^{+0.02}_{-0.02}}$} \\
$0.85$ & $0.999$ & \mbox{$6.4^{+0.1}_{-0.1}$} & \mbox{$5.8^{+0.1}_{-0.01}$} & \mbox{$5.8^{+0.1}_{-0.01}$} & \mbox{$6.1^{+0.1}_{-0.02}$} & \mbox{$6.0^{+0.05}_{-0.05}$} & \mbox{$6.0^{+0.001}_{-0.01}$} & \mbox{$5.4^{+0.1}_{-0.1}$} & \mbox{$\bm{3.1^{+0.03}_{-0.1}}$} \\
$0.9$ & $0.999$ & \mbox{$6.4^{+0.1}_{-0.1}$} & \mbox{$5.8^{+0.05}_{-0.02}$} & \mbox{$5.8^{+0.1}_{-0.01}$} & \mbox{$5.9^{+0.1}_{-0.02}$} & \mbox{$6.1^{+0.1}_{-0.03}$} & \mbox{$6.0^{+0.005}_{-0.01}$} & \mbox{$5.1^{+0.1}_{-0.1}$} & \mbox{$\bm{3.1^{+0.01}_{-0.1}}$} \\
$0.95$ & $0.999$ & \mbox{$6.4^{+0.1}_{-0.1}$} & \mbox{$5.8^{+0.03}_{-0.03}$} & \mbox{$5.9^{+0.02}_{-0.04}$} & \mbox{$5.8^{+0.03}_{-0.03}$} & \mbox{$6.2^{+0.1}_{-0.02}$} & \mbox{$5.9^{+0.002}_{-0.01}$} & \mbox{$4.6^{+0.2}_{-0.03}$} & \mbox{$\bm{3.1^{+0.01}_{-0.03}}$} \\
$0.85$ & $0.9999$ & \mbox{$6.4^{+0.1}_{-0.1}$} & \mbox{$6.0^{+0.05}_{-0.02}$} & \mbox{$6.2^{+0.1}_{-0.03}$} & \mbox{$6.1^{+0.1}_{-0.01}$} & \mbox{$5.8^{+0.05}_{-0.04}$} & \mbox{$6.0^{+0.003}_{-0.004}$} & \mbox{$5.5^{+0.04}_{-0.1}$} & \mbox{$\bm{3.1^{+0.03}_{-0.1}}$} \\
$0.9$ & $0.9999$ & \mbox{$6.4^{+0.1}_{-0.1}$} & \mbox{$6.0^{+0.05}_{-0.01}$} & \mbox{$6.2^{+0.1}_{-0.01}$} & \mbox{$6.0^{+0.1}_{-0.01}$} & \mbox{$5.8^{+0.04}_{-0.03}$} & \mbox{$6.0^{+0.001}_{-0.01}$} & \mbox{$5.2^{+0.1}_{-0.1}$} & \mbox{$\bm{3.1^{+0.01}_{-0.1}}$} \\
$0.95$ & $0.9999$ & \mbox{$6.4^{+0.1}_{-0.1}$} & \mbox{$6.2^{+0.03}_{-0.03}$} & \mbox{$6.3^{+0.1}_{-0.04}$} & \mbox{$6.1^{+0.02}_{-0.03}$} & \mbox{$5.9^{+0.1}_{-0.01}$} & \mbox{$5.9^{+0.003}_{-0.004}$} & \mbox{$4.8^{+0.1}_{-0.05}$} & \mbox{$\bm{3.1^{+0.03}_{-0.03}}$} \\
\bottomrule
\end{tabular}
\end{adjustbox}
\end{table*}

\FloatBarrier

\FloatBarrier

\subsection{TEST 2: univariate scalar complex holomorphic function}\label{sec:test_case_holomorphic}
\paragraph{Rationale}
This relatively inexpensive test case allows us to perform multiple training runs. We consider an important class of complex-valued functions: holomorphic functions, which have applications in, e.g., solid mechanics \cite{Calafa2024, Ballini2026b}. We investigate the effect of the mini-batch size and show that excessively small mini-batches can constitute a failure mode for AURA.

\paragraph{Learning task}
Domain sampling and loss construction follow
\Cref{sec:test_case_non_holomorphic}. The activation and target function are specified below.

The domain is the square
$\Omega_\zeta = \{\zeta\in\CC:|\real\zeta| \leq 1,\ |\imag\zeta| \leq 1\}$.
The target is the entire, multiscale function
\begin{align}
	f_{\rm hol}(\zeta) = {}&
	\exp((0.39-0.22\mathrm i)\zeta) \notag\\
	&+0.12\exp(3.5\zeta)
	+0.03\exp((-2.8+3.3\mathrm i)\zeta) \notag\\
	&+0.002\exp((-5.3-4.6\mathrm i)\zeta) \notag\\
	&+0.0015\exp((5.5+5.0\mathrm i)\zeta) \notag\\
	&+0.015\sin(5.5\zeta)
	+0.059\zeta^4-0.040\mathrm i\zeta^5,
	\label{eq:optimizer_holomorphic_target}
\end{align}
whose oblique complex exponentials produce steep variations near three corners
of $\Omega_\zeta$, while the sine and polynomial terms contribute oscillatory
and algebraic structure.

\paragraph{Network architectures}
The networks follow the affine--activation composition defined in
\Cref{sec:test_case_non_holomorphic}, with $n_0=n_L=1$ and the componentwise
entire activation $\varsigma(\zeta)=\exp(\zeta)$.
Since complex affine maps and the exponential function are entire,
the networks are entire, and hence holomorphic in $\zeta$, by construction.

The complete benchmark configuration, including architectures, activation,
dataset sizes, and optimizer hyperparameters, is given in
\Cref{tab:optimizer_settings_holomorphic}.	

\begin{table}[!ht]
\centering
\caption{Configuration of the Holomorphic optimizer benchmark.}
\label{tab:optimizer_settings_holomorphic}
\small
\begin{tabularx}{\linewidth}{@{}>{\raggedright\arraybackslash}p{0.32\linewidth}X@{}}
\toprule
Setting & Value \\
\midrule
Primary architecture & $(1,32,32,32,32,1)$, \numprint{3265} trainable parameters \\
Secondary architecture & $(1,128,128,128,128,128,1)$, \numprint{66433} trainable parameters \\
Hidden activation & $\exp(\zeta)$ \\
Training / test set & \numprint{2500} / $500$ points (test: 20\%) \\
Mini-batch size & $256$ points \\
Updates / seeds & \numprint{12000} / $\{0,\ldots,4\}$ \\
Arithmetic & \texttt{complex64} (32-bit real components) \\
Initialization & beta-scaled Gaussian-fallback rule for entire activations \cite{Calafa2024}, $\beta_\mathrm{init}=0.5$; zero biases \\
Base learning rate $\alpha$ (primary) & $5\times 10^{-5}$ \\
Secondary-architecture learning rate & $5\times 10^{-6}$ \\
Optimizer hyperparameters & default values; see \Cref{tab:optimizer_settings_shared} \\
\bottomrule
\end{tabularx}
\end{table}

\FloatBarrier

\begin{figure*}[!t]
	\centering
	\textbf{Loss: TEST 2 -- holomorphic}\par\vspace{0.3\baselineskip}
	\begin{minipage}[c]{0.02\linewidth}
		\centering
		\rotatebox{90}{Primary architecture}
	\end{minipage}%
	\begin{minipage}[c]{0.98\linewidth}
		\centering
		\begin{minipage}[c]{0.02\linewidth}
			\centering
			\rotatebox{90}{Mini-batch $=32$}
		\end{minipage}%
		\begin{minipage}[c]{0.98\linewidth}
			\centering
			\includegraphics[width=0.80\linewidth,page=1]{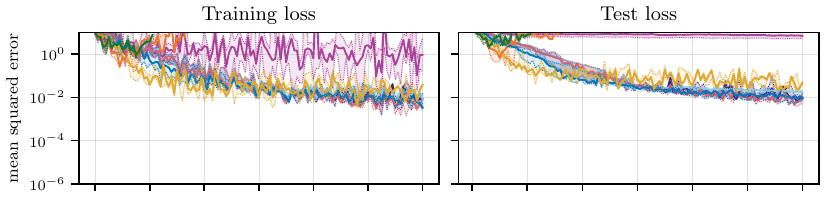}
		\end{minipage}\\
		\begin{minipage}[c]{0.02\linewidth}
			\centering
			\rotatebox{90}{Mini-batch $=128$}
		\end{minipage}%
		\begin{minipage}[c]{0.98\linewidth}
			\centering
			\includegraphics[width=0.80\linewidth,page=2]{figures/optimizer_losses_holomorphic.pdf}
		\end{minipage}\\
		\begin{minipage}[c]{0.02\linewidth}
			\centering
			\rotatebox{90}{Mini-batch $=256$}
		\end{minipage}%
		\begin{minipage}[c]{0.98\linewidth}
			\centering
			\includegraphics[width=0.80\linewidth,page=3]{figures/optimizer_losses_holomorphic.pdf}
		\end{minipage}
	\end{minipage}\\[0.8\baselineskip]
	\begin{minipage}[c]{0.02\linewidth}
		\centering
		\rotatebox{90}{Secondary architecture}
	\end{minipage}%
	\begin{minipage}[c]{0.98\linewidth}
		\centering
		\begin{minipage}[c]{0.02\linewidth}
			\centering
			\rotatebox{90}{Mini-batch $=32$}
		\end{minipage}%
		\begin{minipage}[c]{0.98\linewidth}
			\centering
			\includegraphics[width=0.80\linewidth,page=4]{figures/optimizer_losses_holomorphic.pdf}
		\end{minipage}\\
		\begin{minipage}[c]{0.02\linewidth}
			\centering
			\rotatebox{90}{Mini-batch $=128$}
		\end{minipage}%
		\begin{minipage}[c]{0.98\linewidth}
			\centering
			\includegraphics[width=0.80\linewidth,page=5]{figures/optimizer_losses_holomorphic.pdf}
		\end{minipage}\\
		\begin{minipage}[c]{0.02\linewidth}
			\centering
			\rotatebox{90}{Mini-batch $=256$}
		\end{minipage}%
		\begin{minipage}[c]{0.98\linewidth}
			\centering
			\includegraphics[width=0.80\linewidth,page=6]{figures/optimizer_losses_holomorphic.pdf}
		\end{minipage}
	\end{minipage}
	\par\vspace{0.4\baselineskip}
	\includegraphics{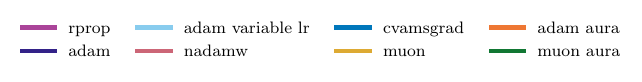}
	\par\vspace{0.2\baselineskip}
	\caption{Training-set (left panel of each row) and test-set (right panel) mean squared errors for the Holomorphic optimizer benchmark. The two architecture blocks show the same feedforward network trained independently at two architectures (\Cref{tab:optimizer_settings_holomorphic}): the primary architecture $(1,32,32,32,32,1)$ (\numprint{3265} parameters) at the top, and the wider and deeper secondary architecture $(1,128,128,128,128,128,1)$ (\numprint{66433} parameters), trained at a ten-times-smaller step size, at the bottom. Within each architecture block, the rows sweep the mini-batch size across $32$, $128$ and $256$ points from top to bottom at a fixed \numprint{12000}-update budget, so a smaller mini-batch means fewer epochs over the training set, not fewer optimizer updates. Solid curves show the pointwise median over all $5$ seeds; shaded regions span the interquartile range (25th--75th percentile) over seeds, whose edges are also drawn as dotted curves.}
	\label{fig:optimizer_losses_holomorphic}
\end{figure*}

\paragraph{Results -- mini-batch size sensitivity}
\Cref{tab:optimizer_training_time_holomorphic} reports $\loss_{\min}^{(m)}$ and $A^{(m)}$ for the mini-batch sizes $32$, $128$, and $256$ at both architectures, together with the number of seeds whose training loss becomes non-finite (NaN); the time ratio is reported only at size $256$, the mini-batch size at which the SGD baseline is measured. Where every seed diverges, $A^{(m)}$ is omitted (--), since an area computed over the steps preceding the divergence is not comparable with that of a complete run. \Cref{fig:optimizer_losses_holomorphic} shows the corresponding training- and test-loss curves.
The update budget and the step size of each architecture are held fixed, so that a smaller mini-batch means fewer passes over the training set, but the same number of updates.
At the largest mini-batch, Muon-AURA and Adam-AURA rank first and second in both $\loss_{\min}^{(m)}$ and $A^{(m)}$ at both architectures, with the minimum loss of Muon-AURA about $22$ to $27$ times lower than that of the best baseline; Muon-AURA remains the best method at size $128$, where Adam-AURA already diverges in one or two seeds, and Muon-AURA and RPROP in one and two seeds at the secondary architecture.
At size $32$ the ranking is reversed: both AURA variants diverge in all five seeds at both architectures.
RPROP, whose step size also follows the agreement between consecutive steps, is the only baseline that diverges, which suggests that the gradient noise of small mini-batches corrupts this agreement signal; Adam, Adam (variable LR), NadamW, CvAMSGrad, and Muon converge in every run.
AURA adds at most $0.13$ to the time ratio of its base optimizer.

\begin{sidewaystable*}
\centering
\caption{Mini-batch-size sensitivity for the Holomorphic benchmark.}
\label{tab:optimizer_training_time_holomorphic}
\small
\setlength{\tabcolsep}{3pt}
\begin{adjustbox}{max width=\linewidth,center}
\begin{tabular}{@{}ll*{10}{c}@{}}
\toprule
 & & \multicolumn{3}{c}{$\mathscr{L}_{\min}^{(m)}$} & \multicolumn{3}{c}{$A^{(m)}$} & \multicolumn{1}{c}{Training time ($\times$SGD)} & \multicolumn{3}{c}{NaN} \\
\cmidrule(lr){3-5}\cmidrule(lr){6-8}\cmidrule(lr){9-9}\cmidrule(lr){10-12}
Arch. & Optimizer & 32 & 128 & 256 & 32 & 128 & 256 & 256 & 32 & 128 & 256 \\
\midrule
\multirow{8}{*}{Primary} & RPROP & \mbox{$(3.3^{+10.1}_{-3.1})\times10^{-2}$} & \mbox{$(1.5^{+7.5}_{-0.9})\times10^{-3}$} & \mbox{$(5.5^{+5.6}_{-1.2})\times10^{-4}$} & \mbox{$10.5^{+0.2}_{-0.5}$} & \mbox{$8.3^{+0.6}_{-0.1}$} & \mbox{$7.81^{+0.09}_{-0.04}$} & \mbox{$1.65^{+0.02}_{-0.02}$} & 1 & 0 & 0 \\
 & Adam & \mbox{$\bm{(3.3^{+5.6}_{-0.4})\times10^{-4}}$} & \mbox{$(2.2^{+2.0}_{-0.04})\times10^{-4}$} & \mbox{$(1.3^{+1.4}_{-0.03})\times10^{-4}$} & \mbox{$8.86^{+0.06}_{-0.03}$} & \mbox{$8.14^{+0.14}_{-0.04}$} & \mbox{$7.8^{+0.2}_{-0.02}$} & \mbox{$\bm{1.55^{+0.03}_{-0.01}}$} & 0 & 0 & 0 \\
 & Adam (variable LR) & \mbox{$(1.2^{+0.2}_{-0.1})\times10^{-3}$} & \mbox{$(6.5^{+1.1}_{-1.9})\times10^{-4}$} & \mbox{$(4.6^{+3.1}_{-3.1})\times10^{-4}$} & \mbox{$8.92^{+0.17}_{-0.03}$} & \mbox{$8.1^{+0.2}_{-0.02}$} & \mbox{$7.8^{+0.3}_{-0.2}$} & \mbox{$1.57^{+0.03}_{-0.03}$} & 0 & 0 & 0 \\
 & NadamW & \mbox{$(3.4^{+1.4}_{-1.4})\times10^{-4}$} & \mbox{$(1.4^{+0.9}_{-0.3})\times10^{-4}$} & \mbox{$(1.2^{+0.7}_{-0.1})\times10^{-4}$} & \mbox{$8.80^{+0.05}_{-0.06}$} & \mbox{$8.0^{+0.1}_{-0.1}$} & \mbox{$7.7^{+0.2}_{-0.01}$} & \mbox{$1.58^{+0.08}_{-0.004}$} & 0 & 0 & 0 \\
 & CvAMSGrad & \mbox{$(6.9^{+5.9}_{-1.3})\times10^{-4}$} & \mbox{$(6.8^{+0.6}_{-3.3})\times10^{-4}$} & \mbox{$(6.1^{+1.2}_{-2.9})\times10^{-4}$} & \mbox{$\bm{8.73^{+0.08}_{-0.001}}$} & \mbox{$8.1^{+0.2}_{-0.1}$} & \mbox{$7.9^{+0.2}_{-0.1}$} & \mbox{$1.63^{+0.04}_{-0.06}$} & 0 & 0 & 0 \\
 & Muon & \mbox{$(7.8^{+4.3}_{-0.4})\times10^{-4}$} & \mbox{$(3.0^{+0.1}_{-0.1})\times10^{-4}$} & \mbox{$(2.1^{+0.3}_{-0.3})\times10^{-4}$} & \mbox{$8.76^{+0.14}_{-0.001}$} & \mbox{$8.06^{+0.11}_{-0.03}$} & \mbox{$7.80^{+0.09}_{-0.01}$} & \mbox{$1.88^{+0.03}_{-0.01}$} & 0 & 0 & 0 \\
 & Adam-AURA & \mbox{$(4.0^{+1.1}_{-2.3})\times10^{-2}$} & \mbox{$(7.1^{+12.2}_{-0.8})\times10^{-5}$} & \mbox{$(5.2^{+0.3}_{-0.8})\times10^{-5}$} & -- & \mbox{$7.7^{+0.1}_{-0.2}$} & \mbox{$6.75^{+0.05}_{-0.03}$} & \mbox{$1.68^{+0.04}_{-0.09}$} & 5 & 2 & 0 \\
 & Muon-AURA & \mbox{$(2.1^{+0.4}_{-1.3})\times10^{-1}$} & \mbox{$\bm{(3.3^{+0.9}_{-0.3})\times10^{-6}}$} & \mbox{$\bm{(5.3^{+2.3}_{-0.4})\times10^{-6}}$} & -- & \mbox{$\bm{5.9^{+0.6}_{-0.003}}$} & \mbox{$\bm{5.72^{+0.06}_{-0.02}}$} & \mbox{$1.90^{+0.01}_{-0.11}$} & 5 & 0 & 0 \\
\cmidrule(lr){1-12}
\multirow{8}{*}{Secondary} & RPROP & \mbox{$(1.3^{+0.2}_{-0.2})\times10^{0}$} & \mbox{$(4.5^{+214637.2}_{-0.4})\times10^{-5}$} & \mbox{$(7.9^{+1.5}_{-0.9})\times10^{-5}$} & -- & \mbox{$7^{+4}_{-0.1}$} & \mbox{$6.97^{+0.02}_{-0.01}$} & \mbox{$1.56^{+0.10}_{-0.03}$} & 5 & 2 & 0 \\
 & Adam & \mbox{$(9.0^{+29.4}_{-4.5})\times10^{-4}$} & \mbox{$(1.8^{+9.3}_{-0.1})\times10^{-4}$} & \mbox{$(1.2^{+2.4}_{-0.2})\times10^{-4}$} & \mbox{$9.4^{+0.02}_{-0.4}$} & \mbox{$8.5^{+0.3}_{-0.1}$} & \mbox{$8.1^{+0.4}_{-0.2}$} & \mbox{$\bm{1.50^{+0.04}_{-0.03}}$} & 0 & 0 & 0 \\
 & Adam (variable LR) & \mbox{$(2.0^{+2.0}_{-1.6})\times10^{-2}$} & \mbox{$(1.3^{+4.5}_{-0.5})\times10^{-3}$} & \mbox{$(2.3^{+42.5}_{-0.02})\times10^{-4}$} & \mbox{$9.6^{+0.4}_{-0.4}$} & \mbox{$8.7^{+0.3}_{-0.2}$} & \mbox{$8.1^{+0.6}_{-0.2}$} & \mbox{$1.52^{+0.03}_{-0.02}$} & 0 & 0 & 0 \\
 & NadamW & \mbox{$(4.2^{+37.8}_{-1.6})\times10^{-4}$} & \mbox{$(9.4^{+59.2}_{-2.2})\times10^{-5}$} & \mbox{$(1.1^{+1.5}_{-0.4})\times10^{-4}$} & \mbox{$9.4^{+0.02}_{-0.4}$} & \mbox{$8.3^{+0.3}_{-0.2}$} & \mbox{$8.0^{+0.4}_{-0.3}$} & \mbox{$1.51^{+0.06}_{-0.005}$} & 0 & 0 & 0 \\
 & CvAMSGrad & \mbox{$(1.1^{+3.4}_{-0.02})\times10^{-3}$} & \mbox{$(5.5^{+20.5}_{-3.8})\times10^{-4}$} & \mbox{$(2.6^{+16.6}_{-0.1})\times10^{-4}$} & \mbox{$9.2^{+0.1}_{-0.2}$} & \mbox{$8.3^{+0.3}_{-0.03}$} & \mbox{$7.9^{+0.5}_{-0.01}$} & \mbox{$1.53^{+0.04}_{-0.01}$} & 0 & 0 & 0 \\
 & Muon & \mbox{$\bm{(8.5^{+0.5}_{-0.8})\times10^{-5}}$} & \mbox{$(6.7^{+1.5}_{-2.4})\times10^{-5}$} & \mbox{$(4.2^{+1.3}_{-0.6})\times10^{-5}$} & \mbox{$\bm{8.8^{+0.3}_{-0.04}}$} & \mbox{$8.4^{+0.2}_{-0.1}$} & \mbox{$8.1^{+0.1}_{-0.1}$} & \mbox{$2.09^{+0.02}_{-0.0004}$} & 0 & 0 & 0 \\
 & Adam-AURA & \mbox{$(4.2^{+1.7}_{-3.5})\times10^{-2}$} & \mbox{$(3.5^{+4.3}_{-0.3})\times10^{-5}$} & \mbox{$(1.9^{+0.6}_{-0.1})\times10^{-5}$} & -- & \mbox{$7.5^{+0.5}_{-0.2}$} & \mbox{$6.67^{+0.06}_{-0.11}$} & \mbox{$1.60^{+0.06}_{-0.06}$} & 5 & 1 & 0 \\
 & Muon-AURA & \mbox{$(1.8^{+0.4}_{-1.1})\times10^{-1}$} & \mbox{$\bm{(7.4^{+20.3}_{-4.8})\times10^{-6}}$} & \mbox{$\bm{(1.6^{+0.1}_{-0.1})\times10^{-6}}$} & -- & \mbox{$\bm{6.7^{+0.4}_{-0.5}}$} & \mbox{$\bm{5.20^{+0.0002}_{-0.05}}$} & \mbox{$2.03^{+0.02}_{-0.02}$} & 5 & 1 & 0 \\
\bottomrule
\end{tabular}
\end{adjustbox}
\end{sidewaystable*}

\FloatBarrier

\FloatBarrier

\subsection{TEST 3: multivariate complex $\CC^4$ function}\label{sec:test_case_multivariate}
\paragraph{Rationale}
This test uses a classical benchmark for complex-valued networks with four complex inputs \cite{Zhang2024, Popa2014, Savitha2009, Savitha2010, Amin2011, Savitha2012}. We investigate the sensitivity of the optimizers to the floating-point precision, training a single architecture in single and double precision, and repeat the step-size sweep of \Cref{sec:test_case_non_holomorphic}.

\paragraph{Learning task}
The target is the multivariate benchmark function considered in
Section~5.1 of \cite{Zhang2024}: 
\begin{equation}
	f_{\rm mc4}(\zeta_1,\zeta_2,\zeta_3,\zeta_4)
	= \frac{1}{1.5}
	\left(\frac{\zeta_2^2}{\zeta_1}+\zeta_3+10\,\zeta_1\zeta_4\right),
	\qquad \boldsymbol\zeta\in\Omega_\zeta.
	\label{eq:optimizer_multivariate_target}
\end{equation}
The training and test sets are drawn as in
\eqref{eq:benchmark_train_set}--\eqref{eq:benchmark_test_set}, with
\begin{equation}
	\Omega_\zeta
	= \bigl\{\boldsymbol\zeta\in\CC^4:
	\real\zeta_k,\imag\zeta_k\in[0.5,1],\ k=1,\ldots,4\bigr\}.
\end{equation}
This domain keeps $\zeta_1$ away from zero, ensuring that the target is
well defined and holomorphic on a neighborhood of $\Omega_\zeta$.
The loss is the mean squared error \eqref{eq:benchmark_batch_loss}, evaluated at the vector-valued inputs $\boldsymbol\zeta$.

\paragraph{Network architecture}
A single network is used, following the affine--activation composition defined in
\Cref{sec:test_case_non_holomorphic} with $n_0=4$ inputs, the components of
$\boldsymbol\zeta=(\zeta_1,\zeta_2,\zeta_3,\zeta_4)\in\CC^4$, and $n_L=1$ output.
The bounded activation
\begin{equation*}
	\varsigma(\zeta)
	= \tanh(\real\zeta)+\mathrm i\tanh(\imag\zeta),
\end{equation*}
is applied componentwise in the hidden layers, so the network is generally non-holomorphic, although the target is holomorphic. Here, unlike in \Cref{sec:test_case_non_holomorphic}, we did not use the $\operatorname{SiLU}$ function, in order to increase the variability of the test cases.

The complete benchmark configuration, including the architecture, activation, dataset sizes, and the swept step sizes and precisions, is given in
\Cref{tab:optimizer_settings_multivariate_c4}.

\begin{table}[!ht]
\centering
\caption{Configuration of the Multivariate C4 optimizer benchmark.}
\label{tab:optimizer_settings_multivariate_c4}
\small
\begin{tabularx}{\linewidth}{@{}>{\raggedright\arraybackslash}p{0.32\linewidth}X@{}}
\toprule
Setting & Value \\
\midrule
Network architecture & $(4,128,128,128,128,128,1)$, \numprint{66817} trainable parameters \\
Hidden activation & $\tanh(\real\zeta)+\mathrm i\tanh(\imag\zeta)$ (componentwise across the four inputs) \\
Training / test set & \numprint{1296} / $259$ points (test: 20\%) \\
Mini-batch size & $128$ points \\
Updates / seeds & \numprint{12000} / $\{0,\ldots,4\}$ \\
Arithmetic & \texttt{complex64} and \texttt{complex128} (single- and double-precision real components; both swept) \\
Initialization & Glorot/Xavier normal rule \cite{Glorot2010}, applied separately to the real and imaginary parts; zero biases \\
Base learning rate $\alpha$ & $1\times 10^{-4}$, swept over $\alpha/10$, $\alpha$ and $10\alpha$ \\
Optimizer hyperparameters & default values; see \Cref{tab:optimizer_settings_shared} \\
\bottomrule
\end{tabularx}
\end{table}

\FloatBarrier

\begin{figure*}[!t]
	\centering
	\textbf{Loss: TEST 3 -- multivariate $\CC^4$}\par\vspace{0.3\baselineskip}
	\begin{minipage}[c]{0.02\linewidth}
		\centering
		\rotatebox{90}{Single precision}
	\end{minipage}%
	\begin{minipage}[c]{0.98\linewidth}
		\centering
		\begin{minipage}[c]{0.02\linewidth}
			\centering
			\rotatebox{90}{$\alpha/10$}
		\end{minipage}%
		\begin{minipage}[c]{0.98\linewidth}
			\centering
			\includegraphics[width=0.80\linewidth,page=1]{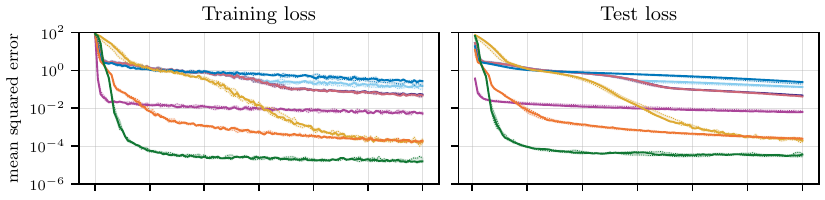}
		\end{minipage}\\
		\begin{minipage}[c]{0.02\linewidth}
			\centering
			\rotatebox{90}{$\alpha$}
		\end{minipage}%
		\begin{minipage}[c]{0.98\linewidth}
			\centering
			\includegraphics[width=0.80\linewidth,page=2]{figures/optimizer_losses_multivariate_c4.pdf}
		\end{minipage}\\
		\begin{minipage}[c]{0.02\linewidth}
			\centering
			\rotatebox{90}{$10\alpha$}
		\end{minipage}%
		\begin{minipage}[c]{0.98\linewidth}
			\centering
			\includegraphics[width=0.80\linewidth,page=3]{figures/optimizer_losses_multivariate_c4.pdf}
		\end{minipage}
	\end{minipage}\\[0.8\baselineskip]
	\begin{minipage}[c]{0.02\linewidth}
		\centering
		\rotatebox{90}{Double precision}
	\end{minipage}%
	\begin{minipage}[c]{0.98\linewidth}
		\centering
		\begin{minipage}[c]{0.02\linewidth}
			\centering
			\rotatebox{90}{$\alpha/10$}
		\end{minipage}%
		\begin{minipage}[c]{0.98\linewidth}
			\centering
			\includegraphics[width=0.80\linewidth,page=4]{figures/optimizer_losses_multivariate_c4.pdf}
		\end{minipage}\\
		\begin{minipage}[c]{0.02\linewidth}
			\centering
			\rotatebox{90}{$\alpha$}
		\end{minipage}%
		\begin{minipage}[c]{0.98\linewidth}
			\centering
			\includegraphics[width=0.80\linewidth,page=5]{figures/optimizer_losses_multivariate_c4.pdf}
		\end{minipage}\\
		\begin{minipage}[c]{0.02\linewidth}
			\centering
			\rotatebox{90}{$10\alpha$}
		\end{minipage}%
		\begin{minipage}[c]{0.98\linewidth}
			\centering
			\includegraphics[width=0.80\linewidth,page=6]{figures/optimizer_losses_multivariate_c4.pdf}
		\end{minipage}
	\end{minipage}
	\par\vspace{0.4\baselineskip}
	\includegraphics{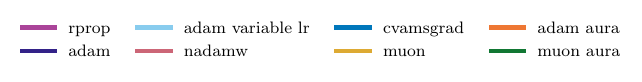}
	\par\vspace{0.2\baselineskip}
	\caption{Training-set and test-set mean squared errors for the Multivariate C4 optimizer benchmark, on the single benchmark architecture $(4,128,128,128,128,128,1)$ (\numprint{66817} parameters; \Cref{tab:optimizer_settings_multivariate_c4}). The two precision blocks show the same network trained in single-precision (\texttt{complex64}, top) and double-precision (\texttt{complex128}, bottom) arithmetic. Within each precision block, the rows sweep the base step size $\alpha = 1\times 10^{-4}$ across $\alpha/10$, $\alpha$ and $10\alpha$ from top to bottom. Solid curves show the pointwise median over all $5$ seeds; shaded regions span the interquartile range (25th--75th percentile) over seeds, whose edges are also drawn as dotted curves.}
	\label{fig:optimizer_losses_multivariate_c4}
\end{figure*}

\paragraph{Results -- $\alpha$ and precision sensitivity}
\Cref{tab:optimizer_training_time_multivariate_c4} reports the three metrics for the step sizes $\alpha/10$, $\alpha$, and $10\alpha$ in single (\texttt{complex64}) and double (\texttt{complex128}) precision, and \Cref{fig:optimizer_losses_multivariate_c4} shows the corresponding training- and test-loss curves.
The precision has little influence: double precision changes $A^{(m)}$ by at most about $0.1$ for every method and step size other than Muon-AURA, and only Muon-AURA improves by more than $0.1$ at every step size, by $0.11$ to $0.28$.
The step size matters far more: over its $100$-fold range, $\loss_{\min}^{(m)}$ varies by more than an order of magnitude for Adam, its variants, and Muon, but by at most about a factor of two for Adam-AURA; RPROP is similarly insensitive, but its minimum loss is about thirty times that of Adam-AURA. 
Muon-AURA attains the lowest $\loss_{\min}^{(m)}$ and $A^{(m)}$ in all six settings, with a margin of about $2$ in $A^{(m)}$ over the best baseline, that is, a geometric-mean training loss about two orders of magnitude lower, and Adam-AURA ranks second in $A^{(m)}$ in all six settings.
AURA adds at most $0.15$ to the time ratio of its base optimizer.

\begin{table*}[!t]
\centering
\caption{Step-size and precision sensitivity for the Multivariate C4 benchmark.}
\label{tab:optimizer_training_time_multivariate_c4}
\small
\setlength{\tabcolsep}{3pt}
\begin{adjustbox}{max width=\linewidth,center}
\begin{tabular}{@{}ll*{7}{c}@{}}
\toprule
 & & \multicolumn{3}{c}{$\mathscr{L}_{\min}^{(m)}\ (\times10^{-4})$} & \multicolumn{3}{c}{$A^{(m)}$} & Training time ($\times$SGD) \\
\cmidrule(lr){3-5}\cmidrule(lr){6-8}\cmidrule(lr){9-9}
Prec. & Optimizer & $\alpha/10$ & $\alpha$ & $10\alpha$ & $\alpha/10$ & $\alpha$ & $10\alpha$ & $\alpha$ \\
\midrule
\multirow{8}{*}{Single} & RPROP & \mbox{$38^{+3}_{-4}$} & \mbox{$36^{+2}_{-1}$} & \mbox{$31^{+5}_{-2}$} & \mbox{$8.04^{+0.01}_{-0.07}$} & \mbox{$7.97^{+0.15}_{-0.01}$} & \mbox{$7.927^{+0.009}_{-0.007}$} & \mbox{$1.55^{+0.05}_{-0.02}$} \\
 & Adam & \mbox{$361^{+19}_{-14}$} & \mbox{$4.1^{+0.2}_{-0.2}$} & \mbox{$2.2^{+0.04}_{-0.1}$} & \mbox{$9.50^{+0.002}_{-0.05}$} & \mbox{$8.02^{+0.02}_{-0.05}$} & \mbox{$7.85^{+0.02}_{-0.01}$} & \mbox{$1.55^{+0.0004}_{-0.11}$} \\
 & Adam (variable LR) & \mbox{$997^{+49}_{-143}$} & \mbox{$12^{+0.5}_{-0.5}$} & \mbox{$1.7^{+0.1}_{-0.1}$} & \mbox{$9.68^{+0.01}_{-0.07}$} & \mbox{$8.15^{+0.005}_{-0.10}$} & \mbox{$7.40^{+0.04}_{-0.002}$} & \mbox{$1.53^{+0.03}_{-0.01}$} \\
 & NadamW & \mbox{$321^{+35}_{-7}$} & \mbox{$7.4^{+0.3}_{-0.4}$} & \mbox{$8.9^{+0.6}_{-3.3}$} & \mbox{$9.50^{+0.001}_{-0.05}$} & \mbox{$8.23^{+0.01}_{-0.04}$} & \mbox{$8.43^{+0.01}_{-0.03}$} & \mbox{$\bm{1.51^{+0.03}_{-0.01}}$} \\
 & CvAMSGrad & \mbox{$1821^{+3}_{-508}$} & \mbox{$18^{+0.1}_{-5}$} & \mbox{$2.2^{+0.2}_{-0.01}$} & \mbox{$9.84^{+0.02}_{-0.06}$} & \mbox{$8.25^{+0.03}_{-0.13}$} & \mbox{$7.34^{+0.03}_{-0.03}$} & \mbox{$1.6^{+0.1}_{-0.1}$} \\
 & Muon & \mbox{$0.8^{+0.01}_{-0.1}$} & \mbox{$27^{+0.5}_{-1}$} & \mbox{$12^{+0.1}_{-0.4}$} & \mbox{$8.35^{+0.03}_{-0.09}$} & \mbox{$7.75^{+0.01}_{-0.02}$} & \mbox{$7.818^{+0.017}_{-0.003}$} & \mbox{$2.128^{+0.013}_{-0.006}$} \\
 & Adam-AURA & \mbox{$1.2^{+0.03}_{-0.04}$} & \mbox{$0.8^{+0.1}_{-0.1}$} & \mbox{$0.9^{+0.2}_{-0.01}$} & \mbox{$7.10^{+0.06}_{-0.01}$} & \mbox{$6.74^{+0.06}_{-0.004}$} & \mbox{$6.68^{+0.09}_{-0.02}$} & \mbox{$1.62^{+0.07}_{-0.002}$} \\
 & Muon-AURA & \mbox{$\bm{0.1^{+0.01}_{-0.003}}$} & \mbox{$\bm{0.1^{+0.01}_{-0.001}}$} & \mbox{$\bm{0.1^{+0.005}_{-0.002}}$} & \mbox{$\bm{5.75^{+0.01}_{-0.03}}$} & \mbox{$\bm{5.64^{+0.01}_{-0.06}}$} & \mbox{$\bm{5.34^{+0.02}_{-0.03}}$} & \mbox{$2.07^{+0.02}_{-0.02}$} \\
\cmidrule(lr){1-9}
\multirow{8}{*}{Double} & RPROP & \mbox{$41^{+23}_{-10}$} & \mbox{$27^{+1}_{-1}$} & \mbox{$32^{+1}_{-5}$} & \mbox{$8.1^{+0.2}_{-0.2}$} & \mbox{$7.85^{+0.04}_{-0.01}$} & \mbox{$7.903^{+0.004}_{-0.005}$} & \mbox{$\bm{1.55^{+0.06}_{-0.01}}$} \\
 & Adam & \mbox{$332^{+42}_{-10}$} & \mbox{$4.1^{+0.1}_{-0.1}$} & \mbox{$2.0^{+0.1}_{-0.1}$} & \mbox{$9.47^{+0.04}_{-0.01}$} & \mbox{$8.01^{+0.01}_{-0.05}$} & \mbox{$7.860^{+0.006}_{-0.010}$} & \mbox{$1.62^{+0.02}_{-0.02}$} \\
 & Adam (variable LR) & \mbox{$1010^{+6}_{-111}$} & \mbox{$10^{+1}_{-0.5}$} & \mbox{$1.7^{+0.2}_{-0.1}$} & \mbox{$9.65^{+0.05}_{-0.02}$} & \mbox{$8.09^{+0.04}_{-0.06}$} & \mbox{$7.411^{+0.002}_{-0.017}$} & \mbox{$1.61^{+0.07}_{-0.07}$} \\
 & NadamW & \mbox{$291^{+46}_{-11}$} & \mbox{$7.0^{+1.1}_{-0.3}$} & \mbox{$7.7^{+1.8}_{-1.4}$} & \mbox{$9.46^{+0.05}_{-0.01}$} & \mbox{$8.22^{+0.005}_{-0.03}$} & \mbox{$8.426^{+0.003}_{-0.010}$} & \mbox{$1.644^{+0.006}_{-0.007}$} \\
 & CvAMSGrad & \mbox{$1540^{+179}_{-170}$} & \mbox{$16^{+4}_{-0.4}$} & \mbox{$2.4^{+0.3}_{-0.1}$} & \mbox{$9.79^{+0.05}_{-0.002}$} & \mbox{$8.16^{+0.08}_{-0.05}$} & \mbox{$7.35^{+0.04}_{-0.02}$} & \mbox{$1.55^{+0.10}_{-0.05}$} \\
 & Muon & \mbox{$1.2^{+0.1}_{-0.1}$} & \mbox{$30^{+0.4}_{-1}$} & \mbox{$14^{+0.4}_{-1}$} & \mbox{$8.4^{+0.2}_{-0.1}$} & \mbox{$7.781^{+0.016}_{-0.0001}$} & \mbox{$7.86^{+0.02}_{-0.05}$} & \mbox{$2.47^{+0.04}_{-0.02}$} \\
 & Adam-AURA & \mbox{$1.2^{+0.2}_{-0.02}$} & \mbox{$0.9^{+0.1}_{-0.1}$} & \mbox{$0.9^{+0.2}_{-0.1}$} & \mbox{$7.14^{+0.02}_{-0.02}$} & \mbox{$6.78^{+0.03}_{-0.02}$} & \mbox{$6.69^{+0.05}_{-0.07}$} & \mbox{$1.76^{+0.06}_{-0.08}$} \\
 & Muon-AURA & \mbox{$\bm{0.1^{+0.01}_{-0.002}}$} & \mbox{$\bm{0.03^{+0.002}_{-0.002}}$} & \mbox{$\bm{0.03^{+0.002}_{-0.001}}$} & \mbox{$\bm{5.64^{+0.05}_{-0.07}}$} & \mbox{$\bm{5.36^{+0.03}_{-0.02}}$} & \mbox{$\bm{5.13^{+0.02}_{-0.02}}$} & \mbox{$2.44^{+0.07}_{-0.04}$} \\
\bottomrule
\end{tabular}
\end{adjustbox}
\end{table*}

\FloatBarrier

\FloatBarrier

\subsection{TEST 4: PINN}\label{sec:test_case_pinn}
\paragraph{Rationale}
This test evaluates the optimizers on a physics-informed learning task while retaining a fully connected holomorphic architecture.
In addition, the hyperparameters of every optimizer are tuned under a limited budget, as recommended in \cite{Schmidt2021}, so that the methods are compared at, or close to, their best performance and no baseline is penalized by an unfavorable default.

\paragraph{Learning task}
The ultimate goal is to solve the two-dimensional linear elasticity problem for a homogeneous material. We consider the geometry and boundary conditions of the plate-with-a-circular-hole benchmark presented in \cite{Calafa2024}. The computational domain is the upper-left quadrant
\begin{equation*}
	\Omega_\zeta=\{\zeta=x+\mathrm i y:\,-a<x<0,\ 0<y<a,\ x^2+y^2>r^2\},
\end{equation*}
with $a=2.5$~m and $r=1$~m.
Traction conditions are imposed on the outer edges and hole arc, while symmetry conditions are imposed on the cut edges; see Section~4.1.2 of
\cite{Calafa2024} for details.

Two holomorphic neural networks, $\varphi_{\mathcal W}:\CC\to\CC$ and $\psi_{\mathcal W}:\CC\to\CC$, take the spatial coordinate $\zeta\in\Omega_\zeta$ as input and approximate the Kolosov--Muskhelishvili potentials $\varphi$ and $\psi$, respectively. Under plane-strain conditions with $\lambda=\mu=1$~MPa, the Kolosov constant is $\kappa=(\lambda+3\mu)/(\lambda+\mu)=2$, and stresses and displacements follow from
\begin{align}
	\sigma_{xx}&=\real(2\varphi'-\overline\zeta\varphi''-\psi'), \notag\\
	\sigma_{yy}&=\real(2\varphi'+\overline\zeta\varphi''+\psi'), \notag\\
	\sigma_{xy}&=\imag(\overline\zeta\varphi''+\psi'), \label{eq:pihnn_stresses}\\
	u_x+\mathrm i u_y&=\frac{1}{2\mu}\bigl(\kappa\varphi-\zeta\overline{\varphi'}-\overline\psi\bigr). \label{eq:pihnn_displacement}
\end{align}
Primes denote differentiation with respect to $\zeta$, evaluated by holomorphic automatic differentiation. This representation satisfies the governing elasticity equations by construction, so training minimizes only the boundary-condition residuals:
\begin{align}
	\loss_{\rm PIHNN}(\mathcal W)={}&\frac{\alpha_N}{|\mathcal Z_N|}\sum_{\zeta\in\mathcal Z_N}\|\bfsigma_{\mathcal W}(\zeta)\mathbf n-\mathbf t_0(\zeta)\|^2 \notag\\
	&+\frac{\alpha_S}{|\mathcal Z_S|}\sum_{\zeta\in\mathcal Z_S}\bigl(\sigma_{xy,\mathcal W}(\zeta)^2+|\bfu_{\mathcal W}(\zeta)\cdot\mathbf n|^2\bigr).
	\label{eq:pihnn_loss}
\end{align}
Here, $\mathcal Z_N$ and $\mathcal Z_S$ contain the points on the traction and symmetry boundaries, respectively, $\mathbf n$ is the outward unit normal, $\mathbf t_0$ is the prescribed traction, and $\alpha_N,\alpha_S$ are the corresponding boundary-length fractions defined in \cite{Calafa2024}. The boundary points remain fixed throughout training, and every update uses the full training set.

\paragraph{Network architecture}
We use two fully connected networks, $\varphi_{\mathcal W}$ and $\psi_{\mathcal W}$, each with
\begin{equation*}
	(n_0,n_1,n_2,n_3,n_4) = (1,64,64,64,1).
\end{equation*}
The three hidden layers use the componentwise activation $\varsigma(\zeta)=\exp(\zeta)$, and the output layer is affine. Both potentials are entire functions of $\zeta$. The networks are wider than in \cite{Calafa2024}, which uses 10 units per hidden layer. A larger network makes the benchmark more meaningful for the optimizers. The wider networks require a smaller learning rate, $10^{-3}$ instead of $10^{-2}$. Adam (variable LR) uses the learning-rate schedule of \cite{Calafa2024}. 

The complete benchmark configuration is given in \Cref{tab:optimizer_settings_pinn}.

\begin{table}[!ht]
	\centering
	\caption{Configuration of the PINN optimizer benchmark.} 
	\label{tab:optimizer_settings_pinn}
	\small
	\begin{tabularx}{\linewidth}{@{}>{\raggedright\arraybackslash}p{0.32\linewidth}X@{}}
		\toprule
		Setting & Value \\
		\midrule
		Architecture (each of $\varphi,\psi$) & $(1,64,64,64,1)$, \numprint{17026} trainable parameters in total \\
		Hidden activation & $\exp(\zeta)$ \\
		Training / test set & $200$ / $20$ boundary points (full batch, no mini-batching) \\
		Updates / seeds & \numprint{6000} / $\{0,\ldots,4\}$ (weight initialization only, boundary sampling fixed) \\
		Arithmetic & \texttt{complex64} (32-bit real components) \\
		Initialization & initialization rule defined in \cite{Calafa2024} (their Algorithm~1), $\beta_\mathrm{init}=0.5$; zero biases \\
		Base learning rate $\alpha$ & $1\times 10^{-3}$, starting point of the search; tuned values in \Cref{tab:optimizer_hpo_pinn} \\
		Optimizer hyperparameters & tuned per method (\Cref{tab:optimizer_hpo_pinn}), starting from the default values of \Cref{tab:optimizer_settings_shared} \\
		\bottomrule
	\end{tabularx}
\end{table}

\FloatBarrier

\paragraph{Hyperparameter optimization}
The learning rate and the principal hyperparameters of every optimizer are searched jointly with the multivariate tree-structured Parzen estimator (TPE) \cite{Bergstra2011} of Optuna \cite{Optuna2026}; \Cref{tab:optimizer_hpo_pinn} lists the search space and the selected values, and every other hyperparameter keeps the value of \Cref{tab:optimizer_settings_shared}.
Every searched quantity is sampled on a logarithmic scale except $\chi_\mathrm{a}$; $\chi_\mathrm{o}$ and $\Psi_\mathrm{a}$ are obtained from the sampled gap $\chi_\mathrm{a}-\chi_\mathrm{o}$ and ratio $\Psi_\mathrm{a}/\Psi_\mathrm{o}$, which preserve the orderings of \Cref{tab:aura_hyperparameters}.
Each method receives the same budget of $50$ trials: the first evaluates the starting point, namely the values of \Cref{tab:optimizer_settings_shared} with $\alpha = 10^{-3}$, sampling is random until ten trials are complete, and TPE proposes the remaining ones.
A trial trains the networks from three weight-initialization seeds, $\{100,101,102\}$, disjoint from the evaluation seeds of \Cref{tab:optimizer_settings_pinn}, and is scored by the mean of $A^{(m,s)}$ \eqref{eq:benchmark_naulc} over them; a trial whose loss becomes non-finite on any seed receives a score of at least $C = 10$, the area of a loss held at $1$, and is excluded from the selection.
The admissible trial with the lowest score is selected and evaluated on the five seeds of \Cref{tab:optimizer_settings_pinn} as in the other cases, without an intermediate confirmation stage; the search costs $8 \times 50 \times 3 = \numprint{1200}$ training runs.
Adam-AURA is searched after Adam, with $\beta_1$ and $\beta_2$ fixed at the values selected for Adam, so that only $\alpha$ and the five gates $\beta_\zeta$, $\chi_\mathrm{a}$, $\chi_\mathrm{o}$, $\Psi_\mathrm{a}$, and $\Psi_\mathrm{o}$ are searched.
This exploits the plug-in property of AURA and isolates the effect of its gates on a tuned base optimizer; counted as a whole, Adam-AURA thus receives twice the budget of the other methods, but the coupling between the Adam and the AURA hyperparameters is left unexplored. Muon-AURA is instead searched independently.

\begin{table*}[!t]
	\centering
	\caption{Hyperparameter search of the PINN benchmark.}
	\label{tab:optimizer_hpo_pinn}
	\small
	\begin{tabularx}{\linewidth}{@{}l>{\raggedright\arraybackslash}X>{\raggedright\arraybackslash}X@{}}
		\toprule
		Optimizer & Search space & Selected values \\
		\midrule
		RPROP & $\alpha\in[10^{-4},10^{-1}]$, $1-\eta_-\in[0.01,0.6]$, $\eta_+-1\in[0.005,0.5]$ & $\alpha=9.727\times 10^{-3}$, $\eta_-=0.7131$, $\eta_+=1.179$ \\
		Adam & $\alpha\in[10^{-4},10^{-1}]$, $1-\beta_1\in[10^{-3},0.3]$, $1-\beta_2\in[10^{-4},0.1]$ & $\alpha=1.888\times 10^{-3}$, $\beta_1=0.9843$, $\beta_2=0.9998$ \\
		Adam (variable LR) & as Adam & $\alpha=3.968\times 10^{-3}$, $\beta_1=0.9851$, $\beta_2=0.9229$ \\
		NadamW & as Adam, and weight decay $\in[10^{-5},0.1]$ & $\alpha=1.249\times 10^{-3}$, $\beta_1=0.9969$, $\beta_2=0.9795$, weight decay $3.437\times 10^{-5}$ \\
		CvAMSGrad & as Adam & $\alpha=3.403\times 10^{-3}$, $\beta_1=0.9818$, $\beta_2=0.998$ \\
		Muon & $\alpha\in[10^{-4},10^{-1}]$, $1-\beta\in[10^{-3},0.3]$ & $\alpha=2.898\times 10^{-4}$, $\beta=0.9962$ \\
		Adam-AURA & $\alpha\in[10^{-4},10^{-1}]$, $1-\beta_\zeta\in[0.01,0.6]$, $\chi_\mathrm{a}\in[0.5,0.98]$, $\chi_\mathrm{a}-\chi_\mathrm{o}\in[0.02,1.2]$, $\Psi_\mathrm{o}\in[0.01,0.9]$, $\Psi_\mathrm{a}/\Psi_\mathrm{o}\in[0.005,0.9]$; $\beta_1$, $\beta_2$ as selected for Adam & $\alpha=1.171\times 10^{-3}$, $\beta_\zeta=0.973$, $\chi_\mathrm{a}=0.7383$, $\chi_\mathrm{o}=-0.4217$, $\Psi_\mathrm{a}=0.02422$, $\Psi_\mathrm{o}=0.03051$ \\
		Muon-AURA & as Muon and as Adam-AURA &  $\alpha=1.816\times 10^{-3}$, $\beta=0.9732$, $\beta_\zeta=0.8016$, $\chi_\mathrm{a}=0.677$, $\chi_\mathrm{o}=-0.0416$, $\Psi_\mathrm{a}=0.0167$, $\Psi_\mathrm{o}=0.1191$ \\
		\bottomrule
	\end{tabularx}
\end{table*}
\FloatBarrier

\begin{figure*}[!t]
	\centering
	\textbf{Loss: TEST 4 -- PINN}\par\vspace{0.3\baselineskip}
	\begin{minipage}[c]{0.02\linewidth}
		\centering
		\rotatebox{90}{Default hyperparameters}
	\end{minipage}%
	\begin{minipage}[c]{0.98\linewidth}
		\centering
		\includegraphics[width=\linewidth]{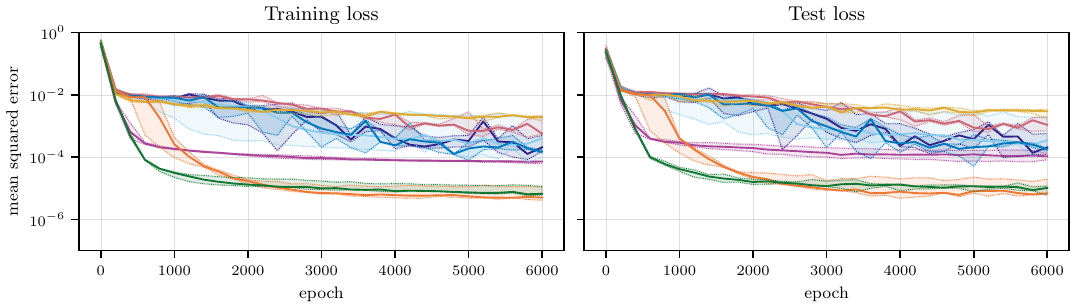}
	\end{minipage}\\[0.4\baselineskip]
	\begin{minipage}[c]{0.02\linewidth}
		\centering
		\rotatebox{90}{Tuned hyperparameters}
	\end{minipage}%
	\begin{minipage}[c]{0.98\linewidth}
		\centering
		\includegraphics[width=\linewidth]{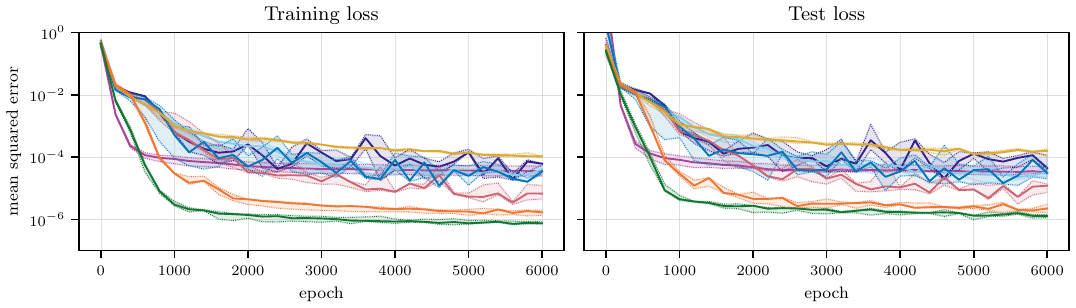}
	\end{minipage}
	\par\vspace{0.4\baselineskip}
	\includegraphics{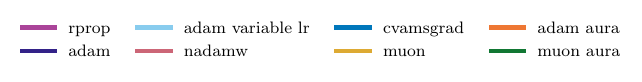}
	\par\vspace{0.2\baselineskip}
	\caption{Training-set and test-set mean squared errors (boundary loss
		\eqref{eq:pihnn_loss}) for the PINN optimizer benchmark. The two
		rows show the same network trained at the default hyperparameters of
		\Cref{tab:optimizer_settings_shared} with $\alpha = 10^{-3}$ (top) and
		at the tuned hyperparameters of \Cref{tab:optimizer_hpo_pinn}
		(bottom). Solid curves
		show the pointwise median over all $5$ seeds; shaded regions span the
		interquartile range (25th--75th percentile) over seeds, whose edges
		are also drawn as dotted curves.}
	 \label{fig:optimizer_losses_pinn}
\end{figure*}

\paragraph{Results}
\Cref{tab:optimizer_training_time_pinn} reports the three metrics at the tuned hyperparameters, together with the number of seeds whose training loss becomes non-finite (NaN), and \Cref{fig:optimizer_losses_pinn} shows the training- and test-loss curves at the tuned (bottom) hyperparameters. For completeness, the same figure also shows the results at the non-tuned hyperparameters, that is, those in \Cref{tab:optimizer_settings_shared}, to better illustrate the effect of the hyperparameter optimization.
Tuning benefits every method, the baselines more than the AURA variants: it lowers $A^{(m)}$ by $0.2$ to $1.8$ for the baselines and by $0.5$ and $0.8$ for Adam-AURA and Muon-AURA, respectively.
The overall conclusion is unchanged: Muon-AURA and Adam-AURA attain the two lowest values of both $\loss_{\min}^{(m)}$ and $A^{(m)}$; the minimum loss of Muon-AURA is about $3.5$ times below that of the best baseline, NadamW, and its margin of $1.3$ in $A^{(m)}$ corresponds to a geometric-mean training loss about $20$ times lower; the minimum loss of Adam-AURA is about $1.7$ times below that of NadamW and its margin of $0.8$ in $A^{(m)}$ to a loss about six times lower.
The largest effect of AURA is on Muon: alone, it reaches the highest minimum loss of all methods, and its combination with AURA lowers it by more than two orders of magnitude.
At their tuned values, which include a step size four to ten times larger than the starting value of the search, RPROP and Adam (variable LR) diverge on one of the five evaluation seeds, although neither diverged on the three search seeds; every other method converges on every seed.
In this full-batch setting, the methods whose step sizes follow the agreement between consecutive steps, and which diverged at small mini-batches in \Cref{sec:test_case_holomorphic}, rank first and second in $A^{(m)}$ (AURA) or within $0.2$ of the best baseline (RPROP), consistent with the interpretation given there.

\begin{table}[!ht]
	\centering
	\caption{Training loss and training time for the PINN benchmark at the tuned hyperparameters of \Cref{tab:optimizer_hpo_pinn}.}
	\label{tab:optimizer_training_time_pinn}
	\footnotesize
	\setlength{\tabcolsep}{3pt}
	\begin{adjustbox}{max width=\linewidth,center}
	\begin{tabular}{@{}l*{4}{c}@{}}
		\toprule
		Optimizer & $\mathscr{L}_{\min}^{(m)}$ & $A^{(m)}$ & Training time ($\times$SGD) & NaN \\
		\midrule
		RPROP & \mbox{$(4.3^{+2.5}_{-2.0})\times10^{-5}$} & \mbox{$5.9^{+0.1}_{-0.2}$} & \mbox{$1.2^{+0.1}_{-0.1}$} & 1 \\
		Adam & \mbox{$(1.0^{+0.02}_{-0.1})\times10^{-5}$} & \mbox{$6.26^{+0.01}_{-0.01}$} & \mbox{$\bm{1.21^{+0.03}_{-0.003}}$} & 0 \\
		Adam (variable LR) & \mbox{$(1.8^{+0.5}_{-0.4})\times10^{-5}$} & \mbox{$6.3^{+0.1}_{-0.2}$} & \mbox{$1.254^{+0.003}_{-0.009}$} & 1 \\
		NadamW & \mbox{$(2.2^{+0.2}_{-0.03})\times10^{-6}$} & \mbox{$5.76^{+0.02}_{-0.02}$} & \mbox{$1.249^{+0.001}_{-0.006}$} & 0 \\
		CvAMSGrad & \mbox{$(5.7^{+1.5}_{-1.2})\times10^{-6}$} & \mbox{$6.0^{+0.2}_{-0.1}$} & \mbox{$1.37^{+0.02}_{-0.02}$} & 0 \\
		Muon & \mbox{$(8.0^{+0.9}_{-1.1})\times10^{-5}$} & \mbox{$6.60^{+0.05}_{-0.005}$} & \mbox{$1.47^{+0.0004}_{-0.05}$} & 0 \\
		Adam-AURA & \mbox{$(1.3^{+0.4}_{-0.1})\times10^{-6}$} & \mbox{$4.96^{+0.05}_{-0.05}$} & \mbox{$1.39^{+0.04}_{-0.03}$} & 0 \\
		Muon-AURA & \mbox{$\bm{(6.3^{+1.6}_{-0.7})\times10^{-7}}$} & \mbox{$\bm{4.46^{+0.01}_{-0.12}}$} & \mbox{$2.09^{+0.01}_{-0.01}$} & 0 \\
		\bottomrule
	\end{tabular}
	\end{adjustbox}
\end{table}

\FloatBarrier

\FloatBarrier

\section{Conclusions}\label{sec:conclusion}

We have proposed AURA, a per-parameter step-size multiplier for training CVNNs.
AURA acts on top of any first-order optimizer and leaves its update direction unchanged.
It compares consecutive update directions of each complex parameter through a Dice-like measure, whose real part quantifies their alignment and whose imaginary part their signed rotation.
The step is enlarged when the directions agree in length, alignment, and sense of rotation, and reduced when they do not.
No additional gradient evaluation is required.
We do not claim that the results achieved are optimal. Rather, the presented findings indicate that AURA can be adopted to improve the results of its base optimizer.

We combined AURA with Adam and Muon and compared the two variants with RPROP, Adam, NadamW, CvAMSGrad, and Muon on four test cases with fully connected CVNNs, from the approximation of scalar complex functions to a physics-informed problem.
The hyperparameters of AURA were held fixed across cases; in \Cref{sec:test_case_pinn}, every optimizer was tuned under the same budget.
Muon-AURA attained the lowest minimum loss and the lowest area under the log learning curve in every setting with mini-batches of at least $128$ points and in the full-batch case, at both fixed and tuned hyperparameters. 
Adam-AURA ranked second in the area metric in most settings.
AURA also reduced the sensitivity to the step size: over a $100$-fold range of $\alpha$, the minimum loss of Adam-AURA varied by less than an order of magnitude, against more than two for Adam.
Tuning benefited the baselines more than the AURA variants, which nonetheless remained the two best methods.
In the mini-batch cases, AURA added at most $0.15$ to the time ratio of its base optimizer.

From the obtained results, we can conclude that AURA fails when the gradient is noisy.
At the smallest mini-batch of \Cref{sec:test_case_holomorphic}, both variants diverged in every seed, together with RPROP. 
The cause lies in the agreement signal itself: AURA inherits no estimate of the gradient variance and measures the agreement between noisy directions.
The full-batch case of \Cref{sec:test_case_pinn}, where the two variants rank first and second, is consistent with this interpretation.

AURA has three further limitations.
It stores five real numbers per parameter in addition to the state of the base optimizer, against the three of Adam.
It adds a few elementwise operations per step.
It introduces ten hyperparameters, of which the decay rate $\beta_\zeta$ and the four thresholds are the least intuitive to set.
On the other hand, the relatively high number of hyperparameters can be an advantage as it represents a margin for adaptation.

Several developments follow from these results.
The failure mode calls for a signal-to-noise estimate in the gate, so that the multiplier is not increased on the agreement of noisy directions.
The current real-valued step size can be naturally extended to a complex step size, which rotates the update.
For real parameters, $\zeta_{j,t}$ is real and $\Psi_{j,t}$ vanishes, so AURA reduces to a Dice-like test of sign and magnitude; whether the gains carry over to real-valued neural networks is an open question.
Convolutional networks were excluded to delimit the scope of this work and are the natural next test.
Finally, for clear computational cost reasons, we tested the proposed method on relatively small networks; whether the results extend to networks with billions of parameters, where the memory overhead matters most, remains to be assessed.

\section*{Data and software availability}
The code that reproduces all numerical experiments of \Cref{sec:benchmarks}, including the scripts that regenerate the tables and figures, is available at \url{https://github.com/enricoballini/aura_optimizer_tests.git}. The setup has been tested on Ubuntu 24.04.

AURA is distributed as the open-source Python package \texttt{aura-optax} (MIT license), which can be installed with \texttt{pip install aura-optax}; it requires Python~3.10 or later and depends only on JAX and Optax.

The package provides \texttt{adam\_aura} and \texttt{muon\_aura}, which implement Adam-AURA and Muon-AURA, and \texttt{scale\_by\_aura}, which implements the AURA step alone and can be chained after any other update rule.

\noindent For complex-valued parameters, the gradients returned by \texttt{jax.grad} must be conjugated before being passed to \texttt{update}, as for every Optax optimizer.

\section*{Acknowledgment}
This work was supported by the Danish Research Council for Independent Research through the grant no. 2035-00142B ``Network-inspired models to predict the strength of heterogeneous materials''. The authors declare that they have no known competing financial interests or personal relationships that could have appeared to influence the work reported in this paper. 

\bibliographystyle{IEEEtran}
\bibliography{all}

@Article{Glorot2010,
  author   = {Glorot,X. and Bengio,Y.},
  journal  = {Journal of Machine Learning Research},
  title    = {Understanding the difficulty of training deep feedforward neural networks},
  year     = {2010},
  pages    = {249-256},
  volume   = {9},
  comment  = {www.scopus.com},
  language = {English},
}

@Article{Kingma2014,
  author    = {Kingma, Diederik P. and Ba, Jimmy},
  eprint        = {1412.6980},
  archivePrefix = {arXiv},
  title     = {Adam: A Method for Stochastic Optimization},
  year      = {2014},
  copyright = {arXiv.org perpetual, non-exclusive license},
  doi       = {10.48550/ARXIV.1412.6980},
 
}

@Article{Calafa2024,
  author    = {Calafà, Matteo and Hovad, Emil and Engsig-Karup, Allan P. and Andriollo, Tito},
  journal   = {Computer Methods in Applied Mechanics and Engineering},
  title     = {Physics-Informed Holomorphic Neural Networks ({PIHNNs}): Solving {2D} linear elasticity problems},
  year      = {2024},
  issn      = {0045-7825},
  month     = dec,
  pages     = {117406},
  volume    = {432},
  doi       = {10.1016/j.cma.2024.117406},
  publisher = {Elsevier BV},
}

@Article{Zhang2024,
  author    = {Zhang, Ke and Zhang, Huisheng and Wang, Xue},
  journal   = {Engineering Applications of Artificial Intelligence},
  title     = {A hybrid complex spectral conjugate gradient learning algorithm for complex-valued data processing},
  year      = {2024},
  issn      = {0952-1976},
  month     = jul,
  pages     = {108352},
  volume    = {133},
  doi       = {10.1016/j.engappai.2024.108352},
  publisher = {Elsevier BV},
}

@Article{Zhang2019,
  author    = {Zhang, Bingjie and Liu, Yusong and Cao, Jinde and Wu, Shujun and Wang, Jian},
  title     = {Fully complex conjugate gradient-based neural networks using Wirtinger calculus framework: Deterministic convergence and its application},
  doi       = {10.1016/j.neunet.2019.02.011},
  issn      = {0893-6080},
  pages     = {50--64},
  volume    = {115},
  journal   = {Neural Networks},
  month     = jul,
  publisher = {Elsevier BV},
  year      = {2019},
}

@Article{Zhao2023,
  author    = {Zhao, Weijing and Huang, He},
  journal   = {Neurocomputing},
  title     = {Adaptive orthogonal gradient descent algorithm for fully complex-valued neural networks},
  year      = {2023},
  issn      = {0925-2312},
  month     = aug,
  pages     = {126358},
  volume    = {546},
  doi       = {10.1016/j.neucom.2023.126358},
  publisher = {Elsevier BV},
}

@Article{McGreivy2024,
  author    = {McGreivy, Nick and Hakim, Ammar},
  journal   = {Nature Machine Intelligence},
  title     = {Weak baselines and reporting biases lead to overoptimism in machine learning for fluid-related partial differential equations},
  year      = {2024},
  issn      = {2522-5839},
  month     = sep,
  number    = {10},
  pages     = {1256--1269},
  volume    = {6},
  doi       = {10.1038/s42256-024-00897-5},
  publisher = {Springer Science and Business Media LLC},
}

@Software{jax2018github,
  author  = {James Bradbury and Roy Frostig and Peter Hawkins and Matthew James Johnson and Yash Katariya and Chris Leary and Dougal Maclaurin and George Necula and Adam Paszke and Jake Vander{P}las and Skye Wanderman-{M}ilne and Qiao Zhang},
  note    = {Open-source software},
  title   = {{JAX}: composable transformations of {P}ython+{N}um{P}y programs},
  version = {0.3.13},
  year    = {2018},
}

@InProceedings{Sashank2018,
  author    = {Sashank J. Reddi and Satyen Kale and Sanjiv Kumar},
  booktitle = {International Conference on Learning Representations},
  title     = {On the Convergence of {Adam} and Beyond},
  year      = {2018},
}

@Article{Kiyani2025,
  author       = {Kiyani, Elham and Shukla, Khemraj and Urbán, Jorge F. and Darbon, Jérôme and Karniadakis, George Em},
  year          = {2025},
  journal = {Computer Methods in Applied Mechanics and Engineering},
  title        = {Optimizing the optimizer for physics-informed neural networks and Kolmogorov-Arnold networks},
  doi          = {10.1016/j.cma.2025.118308},
  issn         = {0045-7825},
  pages        = {118308},
  volume       = {446},
  publisher    = {Elsevier BV},
}

@Article{Ballini2026b,
  author    = {Ballini, Enrico and Engsig-Karup, Allan Peter and Andriollo, Tito},
  year      = {2026},
  eprint        = {2605.31231},
  archivePrefix = {Computer Methods in Applied Mechanics and Engineering, Accepted, arXiv},
  title     = {A holomorphic neural network framework for 3D boundary value problems governed by harmonic potentials},
  doi       = {10.48550/ARXIV.2605.31231},
  copyright = {arXiv.org perpetual, non-exclusive license},
  publisher = {arXiv},
}

@Article{Mayer2025,
  author       = {Mayer, Kayol S. and Soares, Jonathan A. and Cruz, Ariadne A. and Arantes, Dalton S.},
  year          = {2025},
  journal = {IEEE Transactions on Neural Networks and Learning Systems},
  title        = {Adaptive Learning Rate Methods for Complex-Valued Neural Networks},
  doi          = {10.1109/tnnls.2025.3596513},
  issn         = {2162-2388},
  number       = {12},
  pages        = {20157--20170},
  volume       = {36},
  publisher    = {Institute of Electrical and Electronics Engineers (IEEE)},
}

@Article{Zhang2023,
  author       = {Zhang, Yongliang and Huang, He and Shen, Gangxiang},
  year          = {2023},
  journal = {IEEE Transactions on Neural Networks and Learning Systems},
  title        = {Adaptive CL-BFGS Algorithms for Complex-Valued Neural Networks},
  doi          = {10.1109/tnnls.2021.3135553},
  issn         = {2162-2388},
  number       = {9},
  pages        = {6313--6327},
  volume       = {34},
  publisher    = {Institute of Electrical and Electronics Engineers (IEEE)},
}

@Article{Zhao2024,
  author       = {Zhao, Weijing and Huang, He},
  year          = {2024},
  journal = {Expert Systems with Applications},
  title        = {Adaptive stepsize estimation based accelerated gradient descent algorithm for fully complex-valued neural networks},
  doi          = {10.1016/j.eswa.2023.121166},
  issn         = {0957-4174},
  pages        = {121166},
  volume       = {236},
  publisher    = {Elsevier BV},
}

@Article{Cole2021,
  author       = {Cole, Elizabeth and Cheng, Joseph and Pauly, John and Vasanawala, Shreyas},
  year          = {2021},
  journal = {Magnetic Resonance in Medicine},
  title        = {Analysis of deep complex‐valued convolutional neural networks for MRI reconstruction and phase‐focused applications},
  doi          = {10.1002/mrm.28733},
  issn         = {1522-2594},
  number       = {2},
  pages        = {1093--1109},
  volume       = {86},
  publisher    = {Wiley},
}

@Article{Eckhoff2023,
  author       = {Eckhoff, Marco and Reiher, Markus},
  year          = {2023},
  journal = {Journal of Chemical Theory and Computation},
  title        = {Lifelong Machine Learning Potentials},
  doi          = {10.1021/acs.jctc.3c00279},
  issn         = {1549-9626},
  number       = {12},
  pages        = {3509--3525},
  volume       = {19},
  publisher    = {American Chemical Society (ACS)},
}

@Article{Wang2023a,
  author       = {Wang, Yuelin and Wang, Zhidong and Huang, He},
  year          = {2023},
  journal = {Knowledge-Based Systems},
  title        = {Stochastic adaptive CL-BFGS algorithms for fully complex-valued dendritic neuron model},
  doi          = {10.1016/j.knosys.2023.110788},
  issn         = {0950-7051},
  pages        = {110788},
  volume       = {277},
  publisher    = {Elsevier BV},
}

@Article{Zhang2025a,
  author       = {Zhang, Ke and Zhang, Huisheng},
  year          = {2025},
  journal = {Journal of Applied Mathematics and Computing},
  title        = {An adaptive complex-valued stepsize training scheme for complex-valued neural networks},
  doi          = {10.1007/s12190-025-02679-7},
  issn         = {1865-2085},
  number       = {1},
  volume       = {72},
  publisher    = {Springer Science and Business Media LLC},
}

@InProceedings{Yin2026,
  author    = {Yin, Xirui and Huang, He},
  booktitle = {2026 38th Chinese Control and Decision Conference (CCDC)},
  year      = {2026},
  title     = {Flatness Guided Adaptive Gradient Descent Algorithm for Fully Complex-Valued Neural Networks},
  doi       = {10.1109/ccdc69976.2026.11560398},
  pages     = {5944--5949},
  publisher = {IEEE},
}

@Article{Eckhoff2024,
  author       = {Eckhoff, Marco and Reiher, Markus},
  journal = {Machine Learning: Science and Technology},
  title        = {{CoRe} optimizer: an all-in-one solution for machine learning},
  doi          = {10.1088/2632-2153/ad1f76},
  issn         = {2632-2153},
  number       = {1},
  pages        = {015018},
  volume       = {5},
  publisher    = {IOP Publishing},
  year         = {2024},
}

@Article{Roy2021,
  author    = {Roy, S. K. and Paoletti, M. E. and Haut, J. M. and Dubey, S. R. and Kar, P. and Plaza, A. and Chaudhuri, B. B.},
  year      = {2021},
  eprint        = {2105.10190},
  archivePrefix = {arXiv},
  title     = {{AngularGrad}: A New Optimization Technique for Angular Convergence of Convolutional Neural Networks},
  doi       = {10.48550/ARXIV.2105.10190},
  copyright = {arXiv.org perpetual, non-exclusive license},
  publisher = {arXiv},
}

@Article{Dubey2020,
  author       = {Dubey, Shiv Ram and Chakraborty, Soumendu and Roy, Swalpa Kumar and Mukherjee, Snehasis and Singh, Satish Kumar and Chaudhuri, Bidyut Baran},
  journal = {IEEE Transactions on Neural Networks and Learning Systems},
  title        = {{diffGrad}: An Optimization Method for Convolutional Neural Networks},
  doi          = {10.1109/tnnls.2019.2955777},
  issn         = {2162-2388},
  number       = {11},
  pages        = {4500--4511},
  volume       = {31},
  publisher    = {Institute of Electrical and Electronics Engineers (IEEE)},
  year         = {2020},
}

@Article{Lee2022,
  author       = {Lee, ChiYan and Hasegawa, Hideyuki and Gao, Shangce},
  year          = {2022},
  journal = {IEEE/CAA Journal of Automatica Sinica},
  title        = {Complex-Valued Neural Networks: A Comprehensive Survey},
  doi          = {10.1109/jas.2022.105743},
  issn         = {2329-9274},
  number       = {8},
  pages        = {1406--1426},
  volume       = {9},
  publisher    = {Institute of Electrical and Electronics Engineers (IEEE)},
}

@InProceedings{Loshchilov2018,
  author    = {Ilya Loshchilov and Frank Hutter},
  booktitle = {International Conference on Learning Representations},
  title     = {Decoupled Weight Decay Regularization},
  url       = {https://openreview.net/forum?id=Bkg6RiCqY7},
  year      = {2019},
}

@InProceedings{Riedmiller1993,
  author    = {Riedmiller, M. and Braun, H.},
  booktitle = {IEEE International Conference on Neural Networks},
  title     = {A direct adaptive method for faster backpropagation learning: the RPROP algorithm},
  doi       = {10.1109/icnn.1993.298623},
  pages     = {586--591},
  publisher = {IEEE},
}

@InProceedings{Trabelsi2018,
  author    = {Chiheb Trabelsi and Olexa Bilaniuk and Ying Zhang and Dmitriy Serdyuk and Sandeep Subramanian and Joao Felipe Santos and Soroush Mehri and Negar Rostamzadeh and Yoshua Bengio and Christopher J Pal},
  booktitle = {International Conference on Learning Representations},
  title     = {Deep Complex Networks},
  url       = {https://openreview.net/forum?id=H1T2hmZAb},
  year      = {2018},
}

@InProceedings{Popa2014,
  author    = {Popa, Calin-Adrian},
  booktitle = {2014 16th International Symposium on Symbolic and Numeric Algorithms for Scientific Computing},
  year      = {2014},
  title     = {Enhanced Gradient Descent Algorithms for Complex-Valued Neural Networks},
  doi       = {10.1109/synasc.2014.44},
  pages     = {272--279},
  publisher = {IEEE},
}

@Article{Zhang2016,
  author       = {Zhang, Huisheng and Mandic, Danilo P.},
  year          = {2016},
  journal = {IEEE Transactions on Neural Networks and Learning Systems},
  title        = {Is a Complex-Valued Stepsize Advantageous in Complex-Valued Gradient Learning Algorithms?},
  doi          = {10.1109/tnnls.2015.2494361},
  issn         = {2162-2388},
  number       = {12},
  pages        = {2730--2735},
  volume       = {27},
  publisher    = {Institute of Electrical and Electronics Engineers (IEEE)},
}

@InProceedings{Zhuang2020,
  author    = {Zhuang, Juntang and Tang, Tommy and Ding, Yifan and Tatikonda, Sekhar C and Dvornek, Nicha and Papademetris, Xenophon and Duncan, James},
  booktitle = {Advances in Neural Information Processing Systems},
  title     = {{AdaBelief} Optimizer: Adapting Stepsizes by the Belief in Observed Gradients},
  editor    = {H. Larochelle and M. Ranzato and R. Hadsell and M.F. Balcan and H. Lin},
  pages     = {18795--18806},
  publisher = {Curran Associates, Inc.},
  volume    = {33},
  year      = {2020},
}

@InProceedings{Gunes2018,
  author    = {Atilim Gunes Baydin and Robert Cornish and David Martinez Rubio and Mark Schmidt and Frank Wood},
  booktitle = {International Conference on Learning Representations},
  title     = {Online Learning Rate Adaptation with Hypergradient Descent},
  url       = {https://openreview.net/forum?id=BkrsAzWAb},
  year      = {2018},
}

@Article{Kantsila2004,
  author       = {Kantsila, A. and Lehtokangas, M. and Saarinen, J.},
  year          = {2004},
  journal = {Neurocomputing},
  title        = {Complex RPROP-algorithm for neural network equalization of GSM data bursts},
  doi          = {10.1016/j.neucom.2003.11.007},
  issn         = {0925-2312},
  pages        = {339--360},
  volume       = {61},
  publisher    = {Elsevier BV},
}

@Article{Zhang2020,
  author       = {Zhang, Yongliang and Huang, He},
  year          = {2020},
  journal = {Neural Networks},
  title        = {Adaptive complex-valued stepsize based fast learning of complex-valued neural networks},
  doi          = {10.1016/j.neunet.2020.01.011},
  issn         = {0893-6080},
  pages        = {233--242},
  volume       = {124},
  publisher    = {Elsevier BV},
}

@Article{Egghe2009,
  author       = {Egghe, Leo and Leydesdorff, Loet},
  year          = {2009},
  journal = {Journal of the American Society for Information Science and Technology},
  title        = {The relation between Pearson’s correlation coefficient r and Salton’s cosine measure},
  doi          = {10.1002/asi.21009},
  issn         = {1532-2890},
  number       = {5},
  pages        = {1027--1036},
  volume       = {60},
  publisher    = {Wiley},
}

@Article{Savitha2009,
  author       = {Savitha, R. and Suresh, S. and Sundararajan, N. and Saratchandran, P.},
  year          = {2009},
  journal = {Neurocomputing},
  title        = {A new learning algorithm with logarithmic performance index for complex-valued neural networks},
  doi          = {10.1016/j.neucom.2009.06.004},
  issn         = {0925-2312},
  number       = {16-18},
  pages        = {3771--3781},
  volume       = {72},
  publisher    = {Elsevier BV},
}

@InProceedings{Savitha2010,
  author    = {Savitha, R. and Suresh, S. and Sundararajan, N.},
  booktitle = {The 2010 International Joint Conference on Neural Networks (IJCNN)},
  year      = {2010},
  title     = {A self-regulated learning in Fully Complex-valued Radial Basis Function Networks},
  doi       = {10.1109/ijcnn.2010.5596781},
  pages     = {1--8},
  publisher = {IEEE},
}

@InProceedings{Amin2011,
  author    = {Amin, Md. Faijul and Savitha, Ramasamy and Amin, Muhammad Ilias and Murase, Kazuyuki},
  booktitle = {The 2011 International Joint Conference on Neural Networks},
  year      = {2011},
  title     = {Complex-valued functional link network design by orthogonal least squares method for function approximation problems},
  doi       = {10.1109/ijcnn.2011.6033400},
  pages     = {1489--1496},
  publisher = {IEEE},
}

@Article{Savitha2012,
  author       = {Savitha, R. and Suresh, S. and Sundararajan, N.},
  year          = {2012},
  journal = {Neural Networks},
  title        = {A meta-cognitive learning algorithm for a Fully Complex-valued Relaxation Network},
  doi          = {10.1016/j.neunet.2012.02.015},
  issn         = {0893-6080},
  pages        = {209--218},
  volume       = {32},
  publisher    = {Elsevier BV},
}

@Article{Duchi2011,
  author  = {John Duchi and Elad Hazan and Yoram Singer},
  title   = {Adaptive Subgradient Methods for Online Learning and Stochastic Optimization},
  number  = {61},
  pages   = {2121--2159},
  url     = {http://jmlr.org/papers/v12/duchi11a.html},
  volume  = {12},
  journal = {Journal of Machine Learning Research},
  year    = {2011},
}

@Article{Tong2022,
  author       = {Tong, Qianqian and Liang, Guannan and Bi, Jinbo},
  year          = {2022},
  journal = {Neurocomputing},
  title        = {Calibrating the adaptive learning rate to improve convergence of ADAM},
  doi          = {10.1016/j.neucom.2022.01.014},
  issn         = {0925-2312},
  pages        = {333--356},
  volume       = {481},
  publisher    = {Elsevier BV},
}

@InProceedings{Dozat2016,
  author    = {Dozat, Timothy},
  booktitle = {International Conference on Learning Representations (ICLR), Workshop Track},
  title     = {Incorporating {Nesterov} Momentum into {Adam}},
  url       = {https://openreview.net/pdf?id=OM0jvwB8jIp57ZJjtNEZ},
  year      = {2016},
}

@Article{Dong2021,
  author       = {Dong, Zhongying and Huang, He},
  year          = {2021},
  journal = {Neural Networks},
  title        = {A training algorithm with selectable search direction for complex-valued feedforward neural networks},
  doi          = {10.1016/j.neunet.2021.01.014},
  issn         = {0893-6080},
  pages        = {75--84},
  volume       = {137},
  publisher    = {Elsevier BV},
}

@Article{Jacobs1988,
  author       = {Jacobs, Robert A.},
  year          = {1988},
  journal = {Neural Networks},
  title        = {Increased rates of convergence through learning rate adaptation},
  doi          = {10.1016/0893-6080(88)90003-2},
  issn         = {0893-6080},
  number       = {4},
  pages        = {295--307},
  volume       = {1},
  publisher    = {Elsevier BV},
}

@Article{Kesten1958,
  author       = {Kesten, Harry},
  year          = {1958},
  journal = {The Annals of Mathematical Statistics},
  title        = {Accelerated Stochastic Approximation},
  doi          = {10.1214/aoms/1177706705},
  issn         = {0003-4851},
  number       = {1},
  pages        = {41--59},
  volume       = {29},
  publisher    = {Institute of Mathematical Statistics},
}

@TechReport{BartoSutton1981,
  author      = {Barto, Andrew G. and Sutton, Richard S.},
  institution = {Air Force Wright Aeronautical Laboratories, Avionics Laboratory},
  title       = {Goal Seeking Components for Adaptive Intelligence: An Initial Assessment},
  note        = {Technical Report},
  number      = {AFWAL-TR-81-1070},
  address     = {Wright-Patterson Air Force Base, Ohio},
  month       = apr,
  year        = {1981},
}

@Article{Saridis1970,
  author       = {Saridis, George},
  year          = {1970},
  journal = {IEEE Transactions on Systems Science and Cybernetics},
  title        = {Learning Applied to Successive Approximation Algorithms},
  doi          = {10.1109/tssc.1970.300282},
  issn         = {0536-1567},
  number       = {2},
  pages        = {97--103},
  volume       = {6},
  publisher    = {Institute of Electrical and Electronics Engineers (IEEE)},
}

@Article{Zeiler2012,
  author        = {Zeiler, Matthew D.},
  title         = {{ADADELTA}: An Adaptive Learning Rate Method},
  doi           = {10.48550/ARXIV.1212.5701},
  eprint        = {1212.5701},
  archiveprefix = {arXiv},
  copyright     = {arXiv.org perpetual, non-exclusive license},
  year          = {2012},
}

@Article{Riedmiller1994,
  author       = {Riedmiller, Martin},
  year          = {1994},
  journal = {Computer Standards \& Interfaces},
  title        = {Advanced supervised learning in multi-layer perceptrons — From backpropagation to adaptive learning algorithms},
  doi          = {10.1016/0920-5489(94)90017-5},
  issn         = {0920-5489},
  number       = {3},
  pages        = {265--278},
  volume       = {16},
  publisher    = {Elsevier BV},
}

@Misc{Schmidt2021,
  author = {Robin Marc Schmidt and Frank Schneider and Philipp Hennig},
  title  = {Descending through a Crowded Valley {\textemdash} Benchmarking Deep Learning Optimizers},
  url    = {https://openreview.net/forum?id=k2Om84I9JuX},
  year   = {2021},
}

@Misc{Choi2020,
  author = {Dami Choi and Christopher J. Shallue and Zachary Nado and Jaehoon Lee and Chris J. Maddison and George E. Dahl},
  title  = {On Empirical Comparisons of Optimizers for Deep Learning},
  url    = {https://openreview.net/forum?id=HygrAR4tPS},
  year   = {2020},
}

@InProceedings{Schneider2018,
  author    = {Frank Schneider and Lukas Balles and Philipp Hennig},
  booktitle = {International Conference on Learning Representations},
  title     = {Deep{OBS}: A Deep Learning Optimizer Benchmark Suite},
  url       = {https://openreview.net/forum?id=rJg6ssC5Y7},
  year      = {2019},
}

@InProceedings{Zhang2019a,
  author    = {Zhang, Yun and Hua, Qinglong and Xu, Dan and Li, Hongbo and Mu, HuiLin},
  booktitle = {2019 International Radar Conference (RADAR)},
  year      = {2019},
  title     = {A Complex-Valued Convolutional Neural Network with Different Activation Functions in Polarimetric SAR Image Classification},
  doi       = {10.1109/radar41533.2019.171298},
  pages     = {1--4},
  publisher = {IEEE},
}

@Article{Savitha2012a,
  author       = {Savitha, R. and Suresh, S. and Sundararajan, N.},
  year          = {2012},
  journal = {Neural Computation},
  title        = {Metacognitive Learning in a Fully Complex-Valued Radial Basis Function Neural Network},
  doi          = {10.1162/neco_a_00254},
  issn         = {1530-888X},
  number       = {5},
  pages        = {1297--1328},
  volume       = {24},
  publisher    = {MIT Press},
}

@Misc{Jordan2024,
  author = {Keller Jordan and Yuchen Jin and Vlado Boza and You Jiacheng and Franz Cesista and Laker Newhouse and Jeremy Bernstein},
  title  = {Muon: An optimizer for hidden layers in neural networks},
  url    = {https://kellerjordan.github.io/posts/muon/},
  year   = {2024},
}

@Article{Liu2020,
  author    = {Liu, Jie and Lin, Chen and Li, Chuming and Sheng, Lu and Sun, Ming and Yan, Junjie and Ouyang, Wanli},
  year      = {2020},
  eprint        = {2010.11041},
  archivePrefix = {arXiv},
  title     = {Adaptive Gradient Method with Resilience and Momentum},
  doi       = {10.48550/ARXIV.2010.11041},
  copyright = {arXiv.org perpetual, non-exclusive license},
  publisher = {arXiv},
}

@Article{Malviya2024,
  author    = {Malviya, Pranshu and Mordido, Goncalo and Baratin, Aristide and Harikandeh, Reza Babanezhad and Dziugaite, Gintare Karolina and Pascanu, Razvan and Chandar, Sarath},
  year      = {2024},
  eprint        = {2412.18790},
  archivePrefix = {arXiv},
  title     = {Torque-Aware Momentum},
  doi       = {10.48550/ARXIV.2412.18790},
  copyright = {arXiv.org perpetual, non-exclusive license},
  publisher = {arXiv},
}

@InBook{Lu2023,
  author    = {Lu, Jun},
  booktitle = {Sentiment Analysis and Deep Learning},
  year      = {2023},
  title     = {AdaSmooth: An Adaptive Learning Rate Method Based on Effective Ratio},
  doi       = {10.1007/978-981-19-5443-6_21},
  isbn      = {9789811954436},
  pages     = {273--293},
  publisher = {Springer Nature Singapore},
  issn      = {2194-5365},
}

@Article{Goh2007,
  author       = {Goh, Su Lee and Mandic, Danilo P.},
  journal = {IEEE Transactions on Neural Networks},
  title        = {Stochastic Gradient-Adaptive Complex-Valued Nonlinear Neural Adaptive Filters With a Gradient-Adaptive Step Size},
  doi          = {10.1109/tnn.2007.895828},
  issn         = {1045-9227},
  number       = {5},
  pages        = {1511--1516},
  volume       = {18},
  publisher    = {Institute of Electrical and Electronics Engineers (IEEE)},
  year         = {2007},
}

@InProceedings{Goh2005,
  author    = {Su Lee Goh and Mandic, D.P.},
  booktitle = {Proceedings. (ICASSP ’05). IEEE International Conference on Acoustics, Speech, and Signal Processing, 2005.},
  title     = {A Class of Gradient-Adaptive Step Size Algorithms for Complex-Valued Nonlinear Neural Adaptive Filters},
  doi       = {10.1109/icassp.2005.1416288},
  pages     = {253--256},
  publisher = {IEEE},
  volume    = {5},
  year      = {2005},
}

@InProceedings{Balles2018,
  author    = {Balles, Lukas and Hennig, Philipp},
  booktitle = {Proceedings of the 35th International Conference on Machine Learning},
  title     = {Dissecting Adam: The Sign, Magnitude and Variance of Stochastic Gradients},
  editor    = {Dy, Jennifer and Krause, Andreas},
  pages     = {404--413},
  publisher = {PMLR},
  series    = {Proceedings of Machine Learning Research},
  url       = {https://proceedings.mlr.press/v80/balles18a.html},
  volume    = {80},
  month     = {10--15 Jul},
  year      = {2018},
}

@Article{Almansoori2025,
  author       = {Almansoori, Mahmood K.M. and Telek, Miklos},
  year          = {2025},
  journal = {Machine Learning with Applications},
  title        = {Performance evaluation of Complex-Valued Neural Networks on real and complex-valued classification and reconstruction tasks},
  doi          = {10.1016/j.mlwa.2025.100742},
  issn         = {2666-8270},
  pages        = {100742},
  volume       = {22},
  publisher    = {Elsevier BV},
}

@Article{Rutkowski2026,
  author       = {Rutkowski, Michał and Kowalski, Piotr A.},
  year          = {2026},
  journal = {International Journal of Approximate Reasoning},
  title        = {The impact of optimiser choice on the training dynamics of complex-valued neural networks: A comparative study with real-valued counterparts},
  doi          = {10.1016/j.ijar.2026.109722},
  issn         = {0888-613X},
  pages        = {109722},
  volume       = {197},
  publisher    = {Elsevier BV},
}

@Misc{Hinton2012,
  author = {Geoffrey Hinton and Nitish Srivastava and Kevin Swersky},
  title  = {Neural Networks for Machine Learning, Lecture 6e: rmsprop: Divide the gradient by a running average of its recent magnitude},
  note   = {Lecture slides},
  url    = {https://www.cs.toronto.edu/~tijmen/csc321/slides/lecture_slides_lec6.pdf},
  year   = {2012},
}

@Article{Bottou2018,
  author    = {Bottou, Léon and Curtis, Frank E. and Nocedal, Jorge},
  date      = {2018-01},
  title     = {Optimization Methods for Large-Scale Machine Learning},
  doi       = {10.1137/16m1080173},
  issn      = {1095-7200},
  number    = {2},
  pages     = {223--311},
  volume    = {60},
  journal   = {SIAM Review},
  publisher = {Society for Industrial & Applied Mathematics (SIAM)},
}

@Article{Optuna2026,
  author  = {Yoshihiko Ozaki and Shuhei Watanabe and Toshihiko Yanase},
  title   = {OptunaHub: A Platform for Black-Box Optimization},
  number  = {203},
  pages   = {1--10},
  url     = {http://jmlr.org/papers/v27/25-2424.html},
  volume  = {27},
  journal = {Journal of Machine Learning Research},
  year    = {2026},
}

@InProceedings{Bergstra2011,
  author    = {Bergstra, James and Bardenet, R\'{e}mi and Bengio, Yoshua and K\'{e}gl, Bal\'{a}zs},
  booktitle = {Advances in Neural Information Processing Systems},
  title     = {Algorithms for Hyper-Parameter Optimization},
  editor    = {J. Shawe-Taylor and R. Zemel and P. Bartlett and F. Pereira and K. Weinberger},
  publisher = {Curran Associates, Inc.},
  url       = {https://proceedings.neurips.cc/paper_files/paper/2011/file/86e8f7ab32cfd12577bc2619bc635690-Paper.pdf},
  volume    = {24},
  year      = {2011},
}

\begin{figure}[h]
	\centering
	\includegraphics[width=\linewidth]{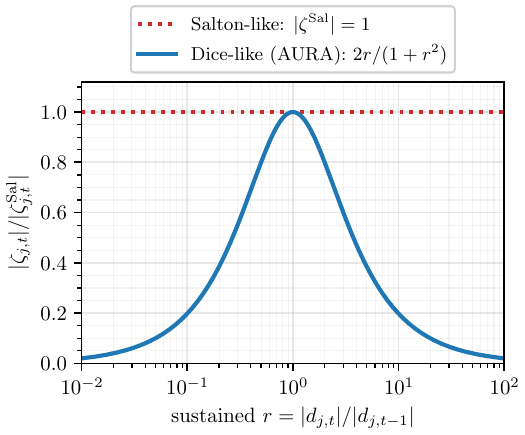}
	\caption{Dice-like against Salton-like measure as a function of the ratio $r = |d_{j,t}|/|d_{j,t-1}|$
	of consecutive step lengths, see \eqref{eq:aura_gm_am}. The Salton-like measure has unit modulus for every
	$r$, whereas the Dice-like one decays as the two lengths separate.}
	\label{fig:aura_gm_am}
\end{figure}

\appendix[Salton-like versus Dice-like measure]
\label{app:salton_dice}
The numerator of $\zeta_{j,t}$ in \eqref{eq:consistency} may be normalized either by the
geometric mean of $|d_{j,t}|^2$ and $|d_{j,t-1}|^2$, that is, $|d_{j,t}||d_{j,t-1}|$, which gives the
Salton-like measure $\zeta^{\rm Sal}_{j,t}$, or by their arithmetic mean, which gives the Dice-like
measure used here. Their real parts are Salton's cosine and Dice's measure of the associated
$\RR^2$ vectors; their imaginary parts have no counterpart in those measures, hence the suffix ``-like''. Writing $r_{j,t} = |d_{j,t}|/|d_{j,t-1}|$ for the ratio of
consecutive step lengths, the two differ by
\begin{equation}\label{eq:aura_gm_am}
	\frac{|\zeta_{j,t}|}{|\zeta^{\mathrm{Sal}}_{j,t}|}
	= \frac{2|d_{j,t}||d_{j,t-1}|}{|d_{j,t}|^2+|d_{j,t-1}|^2}
	= \frac{2r_{j,t}}{1+r_{j,t}^2}\ \in(0,1],
\end{equation}
which equals $1$ only for $r_{j,t} = 1$ and decays as the two lengths separate, see
\Cref{fig:aura_gm_am} and \cite{Egghe2009}. The Dice-like measure therefore damps $\zeta_{j,t}$
whenever consecutive directions differ markedly in magnitude, which is the desired behavior, as it highlights updates that differ in magnitude,
and is used for this reason.

\end{document}